\documentclass[10pt,twocolumn,letterpaper]{article}
\PassOptionsToPackage{hyphens}{url}
\usepackage{latexml}
\iflatexml
\usepackage[numbers,sort&compress]{natbib}
\usepackage{xcolor}
\else
\usepackage[pagenumbers]{objecttwin-preprint}
\fi
\usepackage{amsmath,amssymb}
\usepackage{booktabs}
\usepackage{graphicx}
\usepackage{microtype}
\usepackage{multirow}
\usepackage{tabularx}
\usepackage{xspace}
\newcommand{\method}{ObjectTwin\xspace}
\newcommand{\figref}[1]{Figure~\ref{#1}}
\newcommand{\secref}[1]{Section~\ref{#1}}

\newcommand{\eqnref}[1]{Eq.~\ref{#1}}
\newcommand{\tabref}[1]{Table~\ref{#1}}
\newcommand{\calV}{\mathcal V^{\mathrm{cal}}}
\newcommand{\testV}{\mathcal V^{\mathrm{test}}}
\newcommand{\CD}{\operatorname{CD}}
\newcommand{\IoU}{\operatorname{IoU}}

\definecolor{linkblue}{rgb}{0.21,0.49,0.74}
\usepackage[breaklinks,colorlinks,allcolors=linkblue]{hyperref}
\hypersetup{pdfauthor={Zhenghe Guo},pdftitle={ObjectTwin: Training-Free Asset Verification for Captured Scenes},pdfsubject={Capture-conditioned asset verification}}
\def\confName{Preprint}
\def\confYear{2026}
\title{ObjectTwin: Training-Free Asset Verification for Captured Scenes}
\author{Zhenghe Guo\\
Zhejiang University\\
\texttt{zhenghe.guo@outlook.com}\\[0.4em]
{\small\href{https://styokuok.github.io/sim_ready/}{Project page}\quad
\href{https://arxiv.org/abs/2610.35904}{arXiv:2610.35904}}}
\iflatexml
\else
\AddToHook{cmd/@maketitle/after}{\begin{minipage}{\textwidth}
\centering
\includegraphics[width=\textwidth]{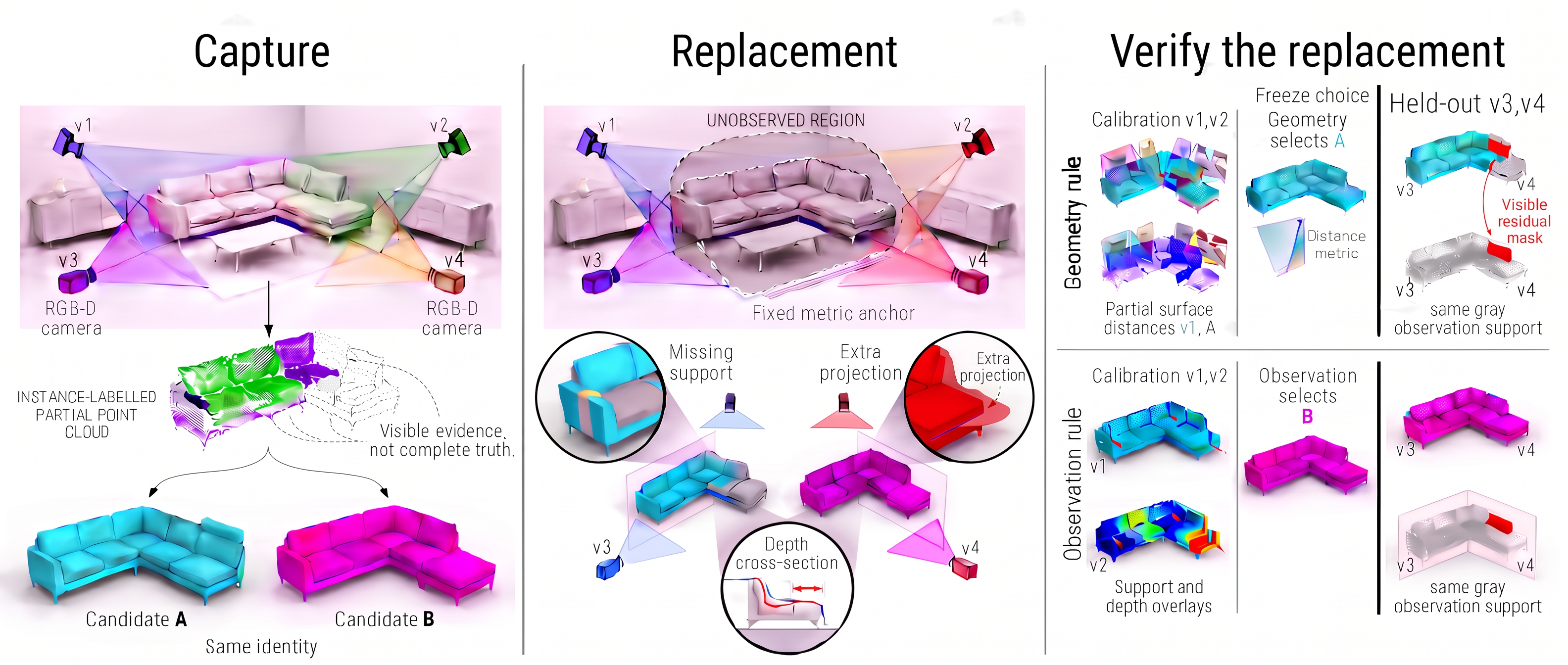}
\captionsetup{hypcap=false}
\captionof{figure}{\textbf{Which asset should represent this captured instance?} \emph{Capture:} posed RGB-D provides partial evidence of the observed object. \emph{Replacement:} shared cameras and placement expose missing support, unsupported projection, and surface-depth disagreement. \emph{Verification:} geometry and observation evidence can favor different candidates. Calibration freezes each choice; held-out views test which asset better explains the captured instance. This conceptual illustration introduces the measured comparison in \figref{fig:selection}.}
\label{fig:teaser}
\end{minipage}
}
\fi
\begin{document}
\raggedbottom
\brokenpenalty=10000
\maketitle
\iflatexml
\begin{figure*}[t]
\centering
\includegraphics[width=\textwidth]{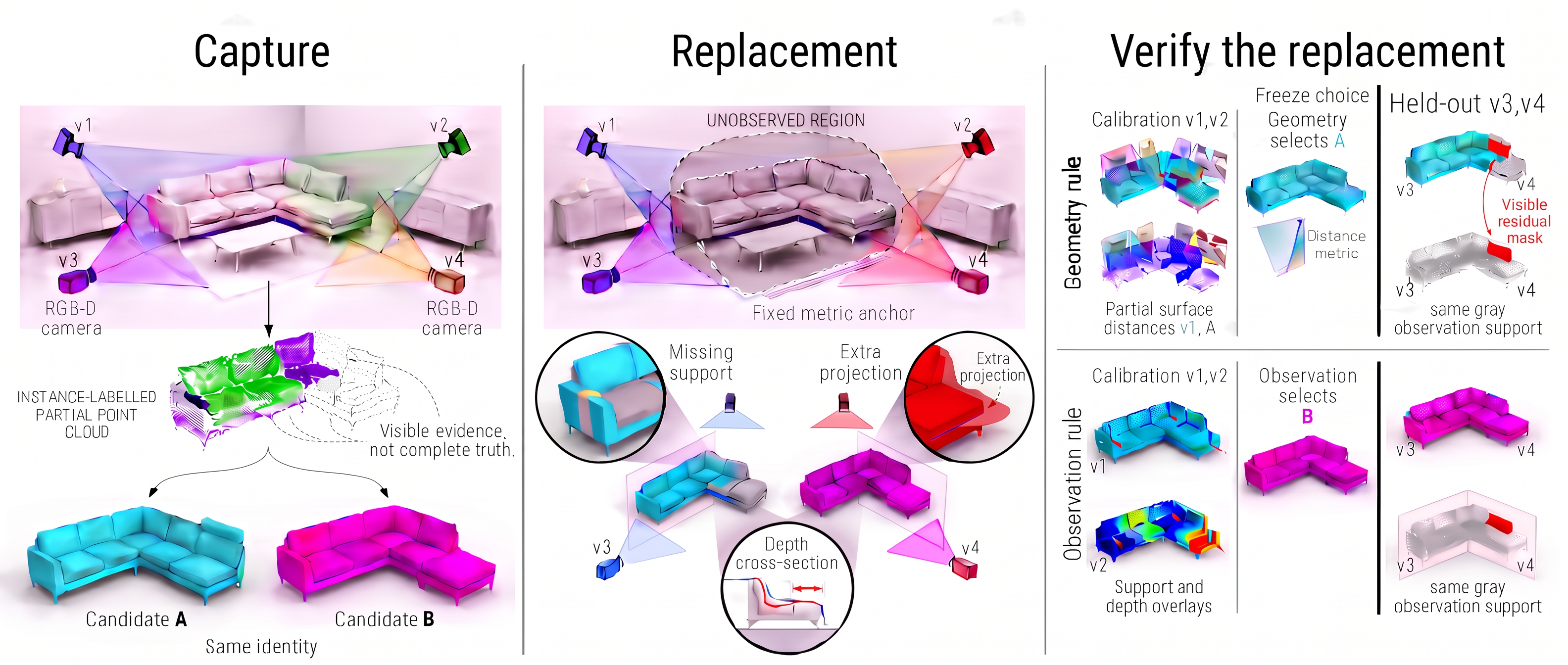}

\caption{\textbf{Which asset should represent this captured instance?} \emph{Capture:} posed RGB-D provides partial evidence of the observed object. \emph{Replacement:} shared cameras and placement expose missing support, unsupported projection, and surface-depth disagreement. \emph{Verification:} geometry and observation evidence can favor different candidates. Calibration freezes each choice; held-out views test which asset better explains the captured instance. This conceptual illustration introduces the measured comparison in \figref{fig:selection}.}
\label{fig:teaser}
\end{figure*}

\fi
\begin{abstract}
An editable digital twin needs assets that faithfully represent the particular objects captured. Partial observations leave both unseen geometry and asset placement uncertain: several registrations can explain the available views while predicting different surfaces elsewhere. We introduce \method, a training-free framework for capture-conditioned asset verification. It retains calibration-compatible alignments, checks observed surfaces and known free space, and carries the resulting range of evidence into acceptance, exclusion, or deferred inspection. Withheld observations evaluate each fixed decision. Across Replica and ScanNet, capture correspondence improves held-out asset selection over registered geometry, reducing mean IoU regret by 55\% on shared candidate pairs. Across multiview reconstruction, amodal completion, and scene generation, surface-based selection identifies the best unseen-view candidate for 97.5\% of shared test objects. Controlled edits verify sensitivity to visible errors and invariance to unobserved geometry. Compatible alignments also predict different unseen outcomes, while strong single-fit controls distinguish this sensitivity from gains in screening accuracy. Together, these experiments establish capture correspondence as a useful asset-selection signal and provide a reproducible framework for evaluating absolute quality and alignment-dependent decisions.
\end{abstract}

\section{Introduction}
An editable digital twin depends on assets that faithfully represent the objects in a captured scene. Recent systems reconstruct objects from casual multiview captures, complete occluded instances, and generate assets together with scene layout~\cite{siddiqui2026shaper,wu2025amodal3r,meng2025scenegen}. These capabilities create a practical evaluation problem: which candidate asset should represent a particular captured object? A visually plausible replacement can alter its projected shape, visible support, and surface depth (\figref{fig:teaser}). Asset selection therefore requires evidence of instance-specific capture correspondence.

Partial capture creates two coupled uncertainties: which surfaces the asset should explain, and where it should be placed. Posed RGB-D constrains visible surfaces and measured free space while leaving occluded geometry unknown. Several registrations can explain these observations similarly, yet predict different surfaces in another view. Scoring only the best registration suppresses this dependence. Selecting the better of two candidates also leaves unanswered whether either satisfies the captured evidence. These difficulties motivate a decision that retains plausible placements and permits unresolved evidence to remain unresolved.

DTC supplies capture-to-scan references~\cite{dong2025digital}; ShapeR evaluates instance geometry across input regimes~\cite{siddiqui2026shaper}; EFM3D measures egocentric surfaces~\cite{straub2024efm3d}. We study the decision following reconstruction: how should capture evidence guide asset choice and acceptance? Shared candidates, registration controls, and unseen test views make this decision measurable.

We introduce \method, a training-free framework for capture-conditioned asset verification. Each candidate retains a finite set of calibration-compatible alignments. Observed-surface and known-free-space discrepancies are evaluated over that set: component requirements test absolute feasibility, and worst-case separation tests whether a relative preference survives the retained placements. The resulting actions are acceptance, exclusion, and deferred inspection. The same decision procedure applies to candidate meshes from different reconstruction routes through explicit capture and coordinate interfaces. It requires no learned evaluator; development requirements and geometric tolerances are fixed before testing.

The experiments connect correspondence, placement ambiguity, and asset decisions. Across ten Replica and ScanNet scenes, observation-based selection improves held-out accuracy from 72.2\% to 86.7\% over registered geometry. External reconstruction experiments extend this finding to multiview object reconstruction, amodal completion, and scene generation. Catalog-based interventions verify detection of visible errors and invariance to changes outside the measured rays. Strong single-fit comparisons then test whether retaining alignment uncertainty improves screening beyond its role in exposing pose-dependent outcomes.

\paragraph{Contributions.}
We formulate capture-conditioned asset verification as a training-free decision over observed evidence and calibration-compatible alignments. We establish the held-out value of capture correspondence through controlled replacement, registration interventions, and shared-candidate selection. Finally, we evaluate alignment dependence and its decision consequences using strong single-fit controls, matched accepted identities, route-specific surface studies, and capture-support audits.

\section{Related Work}
\paragraph{View consistency and selective decisions.}
Ray consistency connects three-dimensional shape to multiview masks and depth~\cite{tulsiani2017ray}. Selective prediction evaluates error together with decision coverage~\cite{geifman2017selective}. \method applies these principles to captured-instance asset verification, where alternative registrations change measured evidence and every candidate may violate a requirement.

\paragraph{Capture-conditioned reconstruction and references.}
DTC provides scanned object models and capture sequences for evaluating object digital twins~\cite{dong2025digital}. ShapeR conditions shape generation on casual captures and compares posed multiview, image-to-3D, and scene-layout methods~\cite{siddiqui2026shaper}. EFM3D evaluates egocentric detection and surface regression~\cite{straub2024efm3d}. These resources offer complementary references, including catalog scans, completed object meshes, and partial capture geometry. \method connects capture-based evaluation to asset choice through controlled replacement, registration analysis, and held-out decision outcomes.

\paragraph{Object and scene generation.}
Amodal3R reconstructs complete objects from occluded images~\cite{wu2025amodal3r}; SceneGen predicts multiple assets and their relative positions from scene images and masks~\cite{meng2025scenegen}. CAST couples component reconstruction with alignment and physics-aware correction~\cite{yao2025cast}. Their evaluation involves different inputs, object-isolation requirements, and placement outputs. Our candidate interface records these differences alongside object identity and transforms. The Tripo/Hunyuan comparison uses paired inputs; the native-system audit retains each route's evidence regime and output conventions.

\paragraph{Alignment and pose.}
Scan2CAD studies CAD-to-scan alignment~\cite{avetisyan2019scan2cad}; FoundationPose and Any6D estimate object pose under different model assumptions~\cite{wen2024foundationpose,lee2025any6d}. \method examines how representation and placement jointly affect capture agreement. Fixed-anchor, calibration-transfer, and free-fit endpoints make registration sensitivity explicit. Archived CAST-orientation and Any6D components provide additional placement controls, with their evaluated scope detailed in the supplement.

\paragraph{Scene fidelity and downstream validation.}
Lucida and GRASE integrate generation and placement into composable or agentic real-to-sim reconstruction~\cite{qin2026lucida,lin2026grase}. RenderBench evaluates render-to-real transfer using reconstructed twins~\cite{qiu2026renderbench}; Splat2Act connects captured environments to manipulation learning~\cite{ye2026splat2act}. These studies show the importance of evaluating reconstructed scenes through their intended use. \method focuses on asset selection within this workflow, testing whether calibration evidence predicts which replacement best explains unseen capture views. Physical interaction introduces further dependencies on collision geometry, material parameters, and policy behavior.

\section{Capture-Conditioned Asset Verification}
\label{sec:protocol}
\begin{figure*}[t]
  \centering
  \includegraphics[width=\textwidth]{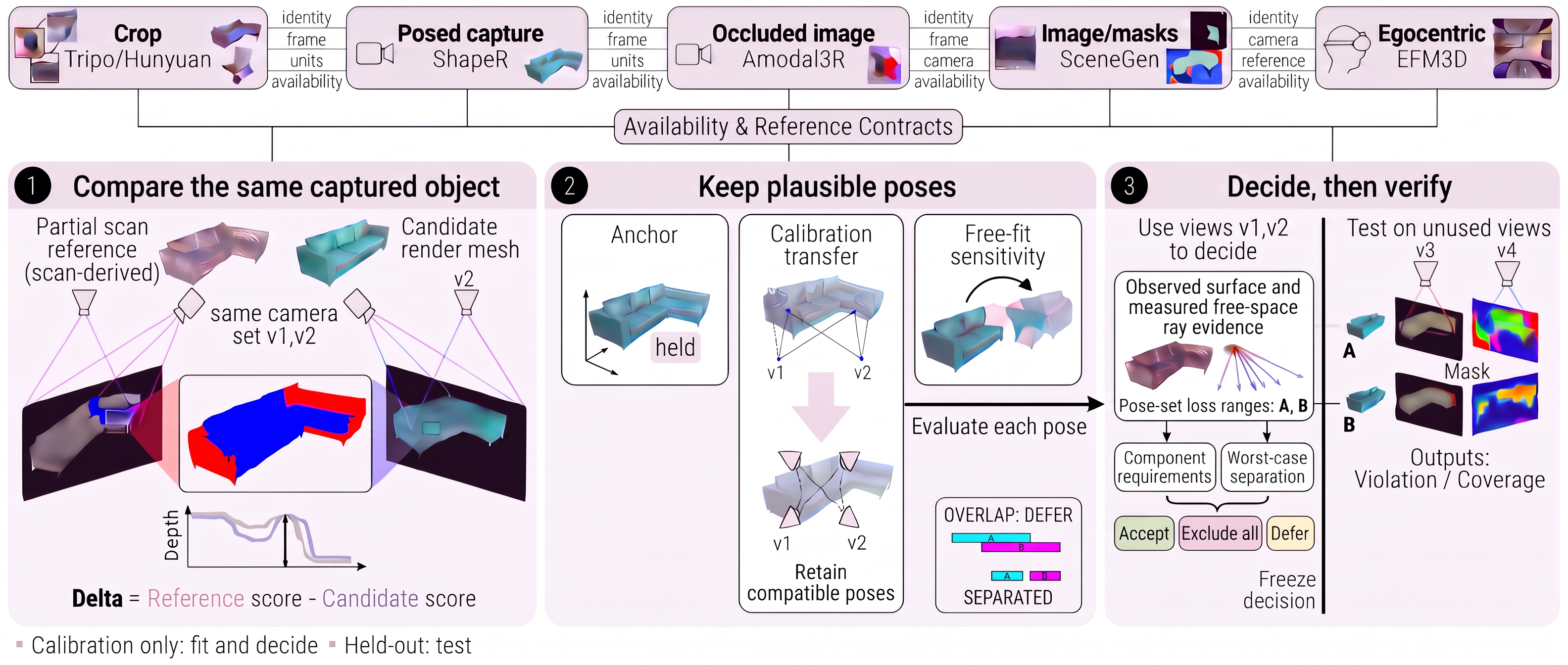}
  \caption{\textbf{Training-free verification from capture evidence to an asset decision.} Paired measurements expose replacement residuals. Calibration transfer retains compatible alignments; anchor and free-fit evaluations supply separate placement controls. Surface and free-space evidence feed component requirements and worst-case separation, producing \emph{Accept}, \emph{Exclude all}, or \emph{Defer}. A candidate passes the requirements; selection resolves the passing alternatives. Decisions freeze before held-out testing. Reconstruction routes enter through identity, coordinate, availability, and reference checks. This conceptual diagram illustrates the interface; measured outcomes appear in \figref{fig:selection} and \tabref{tab:selection}.}
  \label{fig:protocol}
\end{figure*}

\method evaluates candidate meshes against partial capture observations and returns an acceptance, exclusion, or deferred decision (\figref{fig:protocol}). Calibration observations support fitting and decisions; separate held-out observations measure their consequences. We describe the capture interface, geometric evidence, and set-based decision before the registration and ranking controls.
\subsection{Captured evidence and candidate assets}
Each instance supplies posed RGB-D observations, reliable object support, and candidate meshes linked to the same identity and metric frame. Calibration views support fitting and decisions; held-out views are reserved for testing. Unknown support supplies no constraint.

\subsection{Observed surfaces and known free space}

An absolute decision asks whether a candidate explains observed surfaces and respects known free space. For valid object pixels $M_v$ and valid nonobject pixels $N_v$, let $D_v$ and $\widehat D_{mv}(T)$ denote observed and rendered depth. With $\rho(x)=\min(1,x)$, the evidence terms are
\begin{equation}
\begin{aligned}
e_s(u)&=\rho\bigl(|\widehat D_{mv}(u;T)-D_v(u)|/\epsilon\bigr),\quad u\in M_v,\\
e_f(u)&=\rho\bigl([D_v(u)-\widehat D_{mv}(u;T)-\epsilon]_+/\epsilon\bigr),\quad u\in N_v.
\end{aligned}
\label{eq:verification-rays}
\end{equation}
Missing object-ray intersections incur unit surface error. Nonobject rays penalize candidate surfaces in front of an observed surface; unknown rays supply no constraint. Each component is summed and normalized by $|M_v|$, then averaged across calibration views, producing $E_s,E_f$ and $L=E_s+E_f$.

\subsection{Decisions over compatible alignments}
Several placements can explain the available views while predicting different unseen surfaces. Four yaw starts and four spatial-block resamples propose calibration fits. We retain $\mathcal H_m=\{T:L_{\calV}(m,T)\leq\min_{T'}L_{\calV}(m,T')+\delta\}$, with loss interval $[\ell_m,u_m]$. Positive ceilings $b_s,b_f$ specify the component requirements, measured jointly by
\begin{equation}
c_m(T)=\max_{j\in\{s,f\}}\bigl(E_j(m,T)/b_j-1\bigr).
\label{eq:verification-feasibility}
\end{equation}
A candidate passes when $\max_{T\in\mathcal H_m}c_m(T)\leq0$, fails when $\min_{T\in\mathcal H_m}c_m(T)>0$, and otherwise remains unresolved. A missing required measurement also leaves feasibility unresolved. This finite-set criterion retains alignment dependence throughout acceptance.

The decision accepts a uniquely passing candidate when all rivals fail, or a passing candidate satisfying $u_m+\gamma<\ell_n$ for every other passing candidate. It excludes all candidates when each fails. Remaining ambiguity requests inspection within budget and becomes abstention at budget exhaustion. Requirements are frozen on development scenes, and each decision is hashed before opening final-test observations. The supplement specifies settings, support conventions, and stopping outcomes.

For two passing candidates, writing $w_m=u_m-\ell_m$ makes separation equivalent to $\ell_n-\ell_m>w_m+\gamma$. The retained set supplies an alignment-dependent margin: a candidate whose plausible placements vary more needs a larger lead. A fixed-margin single-fit rule therefore isolates the value of this additional information.

\subsection{Replacement and registration controls}

The observations for instance $i$ comprise posed RGB-D frames
$\mathcal O_i=\{I_v,Z_v,K_v,T_v\}$, persistent instance labels, and fused scene geometry.
The evaluation contract fixes the object identity, calibrated cameras, scan-derived visible support, recorded depth, metric anchor $a_i$, and a prediction-independent view set $\mathcal V_i$.
Let $R_{s,i}$ be the instance-labeled scan geometry and $R_{m,i}$ a candidate from method $m$. For an observation score $S$, the replacement gap is
\begin{equation}
 \Delta_{i,m}=S(R_{s,i},a_i;\mathcal O_i,\mathcal V_i)
             -S(R_{m,i},a_i;\mathcal O_i,\mathcal V_i).
 \label{eq:gap}
\end{equation}
The gap measures visible agreement lost under representation replacement at the declared placement. Registration controls below assess its sensitivity to residual placement error.

\paragraph{Candidate interface.}
Each mesh links to an object identity, native input evidence, local-to-capture transform, hash, and availability status. Crop-conditioned generation requires metric placement; scene reconstruction additionally requires object extraction. Comparisons retain each route's input regime and cohort.

\paragraph{Reference construction.}
Projected instance-labeled scan vertices define visible support; recorded depth excludes occluded points. Reference and candidate rendering use the same cameras and occlusion rule. Incomplete geometry and the difference between vertex projection and triangle rendering can reduce reference agreement below one.

\paragraph{Metric anchoring.}
The archived anchor uses scan-derived location, uniform extent-matching scale, and upright PCA yaw. Calibration transfer and free fitting provide separate registration endpoints.

The geometry reference consists of masked observed depth back-projected into world-frame metres. For full candidate surface samples $P$ and partial visible reference points $Q$, our Chamfer convention is
\begin{equation}
\begin{split}
\CD(P,Q)=\frac{1}{2}\bigg[&
 \frac{1}{|P|}\sum_{p\in P}\min_{q\in Q}\|p-q\|_2\\
 &+\frac{1}{|Q|}\sum_{q\in Q}\min_{p\in P}\|q-p\|_2\bigg].
\end{split}
\end{equation}
Distances are unsquared, and F-score uses a 1\,cm threshold. Reference-to-candidate distance measures agreement with observed support; candidate-to-reference distance also includes completed surfaces outside that support. This full-to-partial comparison makes the role of reference visibility explicit.

We sample 10,000 mesh points with seed 0. Modes $a,c,f$ denote anchor, calibration-transfer, and free per-view fitting. Transfer fits one uniform similarity to pooled calibration points using a reference-to-candidate nearest-neighbor objective and deterministic yaw starts. The fitted transform then applies unchanged to held-out views. Free fitting uses target-view geometry exclusively as a sensitivity control. Both optimize a one-sided objective, while the reported bidirectional CD also measures candidate-to-reference distance.

\subsection{Observation profiles and held-out selection}

For reference mask $M_{iv}$ and rendered mask $\widehat M_{imv}$, we compute
\begin{equation}
 S^{\mathrm{IoU}}_{im}=\frac{1}{|\mathcal V_i|}
 \sum_{v\in\mathcal V_i}\frac{|M_{iv}\cap\widehat M_{imv}|}{|M_{iv}\cup\widehat M_{imv}|}.
\end{equation}
Domain aggregates weight instances equally, preventing frame-rich objects from dominating the result. Coverage pools intersection and reference pixel counts within each instance. Overlap-depth L1 pools absolute depth differences on valid intersecting pixels. Together, these diagnostics characterize missing support and surface disagreement where the projections overlap. Empty overlap retains zero support count and undefined depth.

The profile $(S^{\mathrm{IoU}},C,D,z)$ records observation agreement and availability. Full-inventory IoU/coverage assigns zero to missing or invisible renders; conditional quality uses a declared visible subset. Inference success, mesh existence, transform validity, and render visibility retain separate statuses.

For asset selection, the recorded profile rule averages the three observation components. A successful view receives
\begin{equation}
 P_{imv}=\frac{\IoU_{imv}+C_{imv}+\max(0,1-D_{imv}/\tau)}{3},
 \quad \tau=0.05\,\mathrm{m}.
 \label{eq:selector-profile}
\end{equation}
Calibration averages these view scores, assigning $-1$ to failed views. The main choice table scores frozen decisions with bounded held-out IoU, assigning zero to failures. The supplement retains the recorded $-1$-failure utility and component-only controls.

\label{sec:selection}
Each eligible instance has disjoint calibration and held-out sets $\calV_i,\testV_i$. Identifiable generation-input views are assigned to calibration. Geometry selects by minimizing calibration-transfer CD; the profile selects by maximizing \eqnref{eq:selector-profile} at the anchor. Both rules read calibration observations only and require no learned evaluator.

Let $q_{im}$ be a candidate's mean held-out outcome, $\widehat m_i$ the selected candidate, and $q_i^*=\max_m q_{im}$ the retrospective best candidate. We report choice accuracy and regret
\begin{equation}
 \mathrm{Correct}_i=\mathbb 1[q_{i\widehat m_i}=q_i^*],\qquad
 \mathrm{Regret}_i=q_i^*-q_{i\widehat m_i}.
\end{equation}
Predictor ties use lexical candidate ordering, and any oracle-tied choice counts as correct. Random choice uses the analytic uniform expectation over the same candidates. We report both the all-eligible population and a paired ranking cohort with both candidates available at calibration.

\section{Experiments}
\label{sec:experiments}
The experiments ask whether capture evidence changes asset choice, whether plausible alignments lead to different unseen outcomes, and whether retaining those alignments improves acceptance. Dense RGB-D establishes correspondence and decision value; controlled defects and external sparse-ray replay examine the verification mechanism and its transfer.

\subsection{Experimental setup}
The fixed-state study uses five Replica scenes~\cite{straub2019the} and five ScanNet scenes~\cite{dai2017scannet}, comprising 74/63 instances and 370/298 frozen instance-view pairs. Up to five views per instance are selected by scan visibility before inspecting generated outputs.

Selection retains 134 instances and 664 views: 268 calibration and 396 held-out. Three ScanNet instances, totaling four original views, cannot support this split and remain in the fixed-state study. Geometry contains 1,328 candidate-view rows; 1,108 have valid geometry in every registration mode and define the shared comparison set. The remaining 220 retain explicit unavailable status.

\paragraph{Generation and inference.}
Tripo/Hunyuan reuse fixed outputs from identical paired input crops. On the ShapeR multiview cohort~\cite{siddiqui2026shaper}, we compare ShapeR, Amodal3R, and SceneGen~\cite{wu2025amodal3r,meng2025scenegen}; DTC~\cite{dong2025digital} compares ShapeR with EFM3D~\cite{straub2024efm3d}. External methods use pretrained outputs. Registration and direct choice require no evaluator training. A separate pose-feature control uses ridge regression. Input and reference details appear in the supplement.

\paragraph{Uncertainty.}
Replacement contrasts use 4,000 scene--instance bootstrap replicates and leave-one-scene-out (LOO) ranges across five scenes per domain. All-eligible selection uses 10,000 bootstrap replicates with seed 0. Paired selection and zero-failure rescoring report descriptive estimates, each with its own cohort and outcome convention.

\subsection{Capture correspondence and asset choice}
\begin{figure*}[t]
  \centering
  \includegraphics[width=\textwidth]{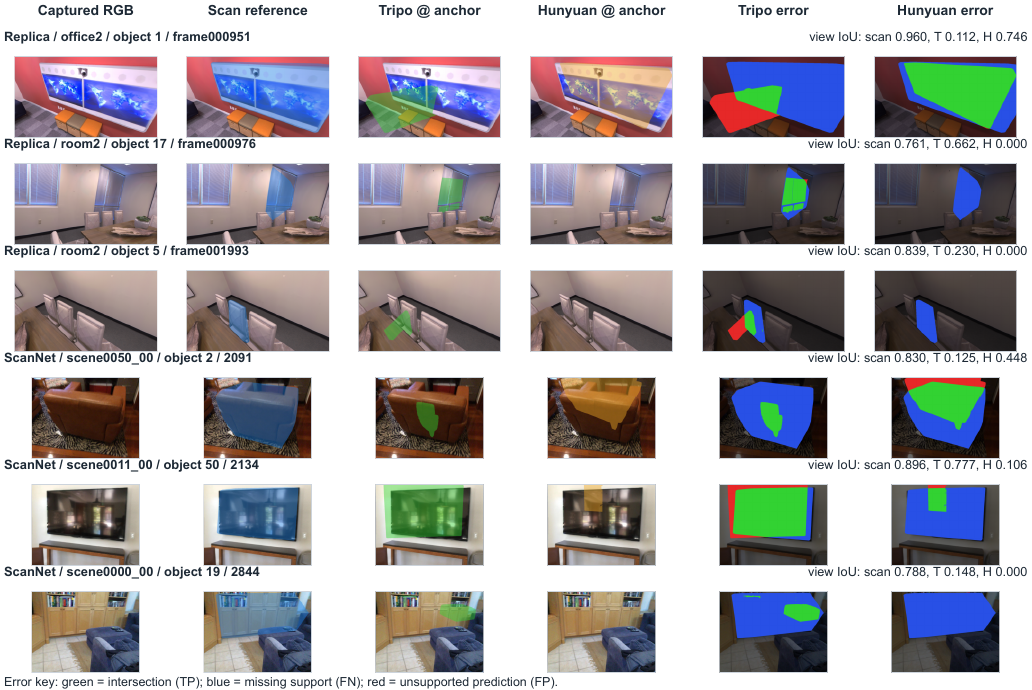}
  \caption{\textbf{One camera, six diagnostic columns, two capture domains.} Each domain includes a Hunyuan-higher case, a Tripo-higher case, and a low--low case with different residuals. All columns in a row share the captured camera and archived anchor; labels report that view's IoU. Green denotes intersection, blue missing reference support, and red unsupported prediction. The supplement expands this fixed-state comparison to all 24 archived cases and presents the view-disjoint selection evidence separately.}
  \label{fig:witnesses}
\end{figure*}
\begin{table*}[t]
  \centering
  \caption{\textbf{Fixed-state representation replacement loses visible agreement.} Instance-macro scan-reference minus generated@anchor MaskIoU, conditioned on a visible anchored proxy. Hunyuan/Replica has 30 visible proxies among 74 requested instances. Confidence intervals and LOO ranges use the same conditional cohort.}
  \label{tab:gap}
  \small\begin{tabular}{llrrll}
\toprule
Domain & Backend & $n$ & Gap & 95\% CI & LOO range \\
\midrule
Replica & Tripo & 74 & 0.358 & [0.294, 0.433] & [0.342, 0.385] \\
Replica & Hunyuan & 30 & 0.302 & [0.221, 0.380] & [0.293, 0.317] \\
ScanNet & Tripo & 63 & 0.310 & [0.222, 0.388] & [0.294, 0.351] \\
ScanNet & Hunyuan & 63 & 0.497 & [0.426, 0.558] & [0.483, 0.519] \\
\bottomrule
\end{tabular}

\end{table*}

Replacement reduces visible agreement for both generators in both capture domains. Conditional MaskIoU gaps range from 0.302 to 0.497 (\tabref{tab:gap}); all bootstrap intervals and LOO ranges exclude zero.

Availability and generator ordering vary. Hunyuan/Replica lacks 44 of 74 GLBs; its missing-as-zero gap is 0.534, versus 0.302 conditionally. On original paired instances, Hunyuan minus Tripo is $+0.110$ on Replica and $-0.186$ on ScanNet, motivating instance-specific choice. Coverage and depth expose distinct residuals (\figref{fig:witnesses}; full results in the supplement).

\paragraph{Registration sensitivity and residual error.}
\begin{table*}[t]
  \centering
  \caption{\textbf{Registration sensitivity on shared valid geometry rows.} CD is the mean unsquared bidirectional distance in centimetres (lower is better); F is F-score in percent at 1\,cm (higher is better). Subscripts $a,c,f$ denote archived anchor, calibration-transfer, and free per-view fit. Each cell averages views within instances, then instances. All modes use the same valid row set within a cell. The reference is partial visible depth; candidate samples cover the full mesh.}
  \label{tab:geometry}
  \small\begin{tabular}{llrrrrrrr}
\toprule
Domain & Candidate & $n$ & CD$_a$ & CD$_c$ & CD$_f$ & F$_a$ & F$_c$ & F$_f$ \\
\midrule
Replica & Hunyuan & 30 & 16.34 & 14.23 & 15.34 & 5.8 & 9.3 & 9.7 \\
Replica & Tripo & 74 & 23.81 & 18.48 & 18.27 & 5.0 & 6.3 & 6.8 \\
ScanNet & Hunyuan & 60 & 28.69 & 24.19 & 24.14 & 1.6 & 2.5 & 2.9 \\
ScanNet & Tripo & 60 & 16.69 & 13.16 & 11.98 & 5.4 & 10.6 & 12.3 \\
\bottomrule
\end{tabular}

\end{table*}

Registration explains part of the measured disagreement. Calibration transfer reduces partial-reference CD in every domain--generator cell (\tabref{tab:geometry}): Replica/Tripo changes from 23.81 to 18.48\,cm and ScanNet/Tripo from 16.69 to 13.16\,cm. Reporting the three registration modes exposes this placement sensitivity alongside the remaining residual.

Transfer also reduces held-out CD in all four cells; the supplement reports these results and free-fit sensitivity. Substantial residual remains after adaptation.

\paragraph{Held-out asset choice.}
The principal choice test compares the same 90 instances with both candidates available at calibration. Calibration evidence selects one asset per instance; held-out views determine correctness and IoU regret, with failed views contributing zero. This shared-candidate comparison tests ranking when both selectors have the same available alternatives.
\begin{figure*}[!t]
  \centering
  \includegraphics[width=\textwidth]{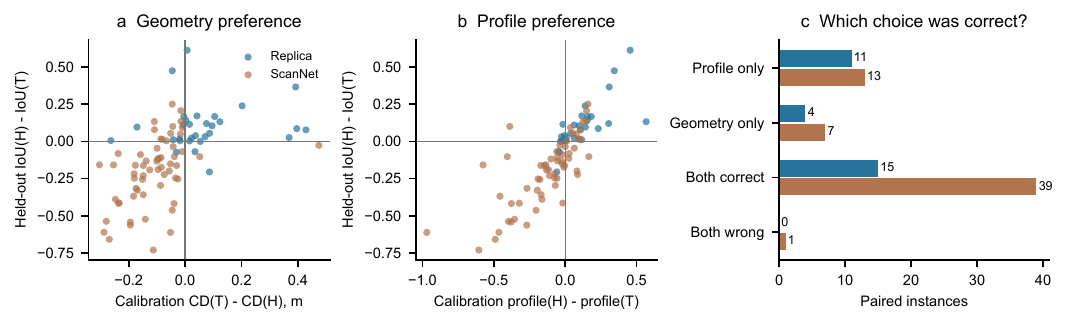}
  \caption{\textbf{Observation evidence resolves asset choices that geometry ranks differently.} Every point is one of the same 90 paired instances. Positive margins favor Hunyuan; geometry's sign is reversed because smaller CD is better. Panels (a,b) compare calibration preference with unseen-view IoU, using zero for failed held-out renders. Same-sign quadrants indicate correct choices. Panel (c) retains both disagreement directions: the profile alone is correct for 24 instances, geometry alone for 11.}
  \label{fig:selection}
\end{figure*}

\begin{table*}[!t]
  \centering
  \caption{\textbf{Observation-based selection improves held-out asset agreement.} Both candidates are available at calibration: 30 Replica and 60 ScanNet instances. Fixed calibration decisions are scored by held-out IoU, assigning zero to failed views before averaging. CD selects by calibration-transfer geometry; the profile uses \eqnref{eq:selector-profile} at the anchor.}
  \label{tab:selection}
  \small\begin{tabular}{lrlrrr}
\toprule
Domain & $n$ & Fixed selector & Correct & Acc.\,(\%) $\uparrow$ & IoU regret $\downarrow$ \\
\midrule
Replica & 30 & Cal.-transfer CD & 19/30 & 63.3 & 0.0386 \\
Replica & 30 & Cal. profile & 26/30 & 86.7 & 0.0062 \\
ScanNet & 60 & Cal.-transfer CD & 46/60 & 76.7 & 0.0210 \\
ScanNet & 60 & Cal. profile & 52/60 & 86.7 & 0.0150 \\
Pooled & 90 & Cal.-transfer CD & 65/90 & 72.2 & 0.0269 \\
Pooled & 90 & Cal. profile & 78/90 & 86.7 & 0.0121 \\
\bottomrule
\end{tabular}

\end{table*}

On the 90-instance paired cohort, the archived profile is correct on 78 instances (86.7\%), compared with 65 (72.2\%) for registered CD (\tabref{tab:selection}). The domain breakdown is 26/30 versus 19/30 on Replica and 52/60 versus 46/60 on ScanNet. Calibration observation agreement therefore identifies a better unseen-view proxy more often than the tested geometry selector.

Mean held-out IoU regret falls by 55\%, from 0.0269 to 0.0121, improving in both domains.

\paragraph{Diagnostic components and selector performance.}
Silhouette agreement is a strong selection signal. Post-hoc calibration-only rules on the same 90 pairs yield 85 correct for IoU, 73 for coverage, and 62 for depth, versus 78 for the profile; IoU-only regret is 0.0024. Coverage achieves 28/30 correct on Replica and 45/60 on ScanNet. Support and depth diagnose different errors, while their predictive value depends on the selection rule (full component results in the supplement).

The net 13 additional correct choices arise from 35 disagreements (\figref{fig:selection}): profile alone is correct on 24 and geometry alone on 11. Both are correct on 54 and wrong on one. Across the ten scenes, profile accuracy improves in seven, ties in one, and decreases in two. The supplement reports scene-level accuracy and regret alongside both directions of decision disagreement.

\paragraph{Failure-policy sensitivity.}
The original $-1$-failure utility preserves the paired ordering. All-eligible accuracy is 91.0\% versus 81.3\% on 134 instances; the supplement reports bootstrap intervals and both failure conventions.

\paragraph{Pose-feature incremental control.}
A separate leave-one-scene-out ridge control adds correspondence features to CAST-orientation/HYBRID observation features. Replica regret falls from 0.0272 to 0.0003; ScanNet retains near-ceiling performance (supplement).

\subsection{Does alignment ambiguity affect unseen evidence?}
Controlled edits first check the geometric evidence. Increasing a visible hole raises surface loss from 0.058 to 0.268; enlarging an added component raises free-space loss from 0.054 to 0.483. An unseen rear addition leaves both unchanged. Calibration translation removes a 0.16\,m placement error while the hole retains its surface discrepancy. These checks separate placement sensitivity, visible representation errors, and unknown completion.

Catalog-based controls on 16 DTC development identities detect every 5\,cm placement error and every 2\,cm local surface defect; all 16 unobserved edits preserve evidence exactly. Fifteen references meet the development requirements. One remains above the surface bound after first-hit visibility filtering; the supplement reports its support association and the full control audit.

On 54 DTC test captures, ShapeR and EFM3D yield 108 candidate alignment sets: 38 retain multiple distinct transforms and 28 have held-out loss ranges above 0.05. Across ShapeR, Amodal3R, and calibration-registered SceneGen, 14 of 372 test alignment sets have surface-loss ranges above 0.05. Compatible placements therefore predict different unseen evidence across reconstruction methods. The supplement details the camera calibration, reference association, and results by method.

\subsection{Does set-based screening improve acceptance?}
The screening controls share candidates, alignment proposals, quality thresholds, and test observations. Geometry uses its calibration-best geometry pose; single-fit screening uses its best ray pose; set-based screening tests feasibility across retained poses; separation additionally requires a stable relative preference. Risk is the fraction of accepted assets violating held-out requirements. Matching accepted identities distinguishes improved screening from lower coverage.

\paragraph{Scene-disjoint dense RGB-D.}
A scene-disjoint verification study reserves three Replica room2 instances for development and evaluates 27 instances from four other scenes. With up to two additional observations, it accepts nine candidates, excludes all candidates for 14 instances, and leaves four unresolved. Seven accepted candidates meet the frozen surface/free-space requirements on held-out observations; two do not.

A common-endpoint replay uses the same calibration-best ray pose per candidate. Forced IoU choice hands off 17 noncompliant candidates; single-fit screening accepts 13 with three violations; envelope plus separation accepts nine with two. On those same nine identities, single-fit screening also has two violations. The supplement reports all component ablations.

\paragraph{External surface choice and screening.}
The DTC comparison uses 16 development and 54 test captures, all active. With nearest-surface reference association, single and set screening make identical decisions: ten accepted assets, eight held-out violations. First-visible-surface filtering is evaluated separately in the reference controls. The supplement specifies each reference construction and its associated quality thresholds.

We compare surface-based selection among ShapeR, Amodal3R, and SceneGen on 122 shared test objects. All selectors evaluate their chosen candidate at its calibration-best surface pose. SceneGen uses calibration registration because its output geometry is normalized. This post-hoc comparison evaluates relative surface quality without an absolute acceptance threshold.
\begin{table}[t]
\centering
\caption{\textbf{Three-route choice on 122 shared objects.} Correctness and regret use held-out surface loss at common deployment poses. SceneGen uses calibration registration; the other routes retain recorded frame initialization.}
\label{tab:external-surface-choice}
\footnotesize\begin{tabular}{lrrr}
\toprule
Rule & Correct & Loss $\downarrow$ & Regret $\downarrow$ \\
\midrule
Fixed route (dev.) & 63 & 0.9379 & 0.01653 \\
Geometry selection & 91 & 0.9263 & 0.00484 \\
Single-fit surface & 119 & 0.9214 & 0.00002 \\
Worst-case surface & 93 & 0.9238 & 0.00241 \\
\bottomrule
\end{tabular}

\end{table}

Single-surface choice identifies the held-out-best candidate for 119/122 objects, versus 91 for geometry, 93 for the upper envelope, and 63 for a fixed route selected on development objects. It also has lower mean regret than envelope ranking (\tabref{tab:external-surface-choice}). These outcomes establish the predictive value of observation evidence across routes; set-based screening supplies no additional gain in these comparisons.

\subsection{Evidence across reconstruction and capture regimes}
The multiview capture study evaluates 125 test objects for ShapeR/Amodal3R and 122 for SceneGen across five scenes. Measured object points and optical centers provide surface reference rays. The supplement reports loss distributions, alignment sensitivity, and per-scene selection results, together with fitting-budget, point-count, and depth-tolerance analyses on 16 DTC development objects.

The DTC results describe active capture, while Replica and ScanNet establish the two-domain dense RGB-D comparison. References are capture-derived or catalog-associated; independent manual-mask validation is not included. Each study reports its reference construction, registration procedure, and evaluation population.

\section{Discussion and Conclusion}
\method carries capture evidence and plausible registrations into explicit asset decisions. Observation-based choice improves held-out agreement over registered geometry in both indoor domains and predicts unseen surface quality across three reconstruction routes. Controlled defects verify sensitivity to visible errors and invariance to unobserved edits. Distinct retained alignments predict different unseen outcomes; strong single-fit controls separate this sensitivity from screening gains. Capture correspondence therefore provides an effective asset-choice signal, with explicit reference and placement checks for absolute decisions.

\label{page:last-main}
\clearpage
\section*{Acknowledgments}
This work uses the Replica and ScanNet datasets. The author acknowledges research funding administered through Shenzhen University and computational resources provided by the State Key Laboratory of CAD\&CG, Zhejiang University.
\section*{LLM Usage Statement}
The author directed the research, conducted the experiments, verified the results, and revised and approved the final manuscript. OpenAI Codex assisted with technical discussion, manuscript preparation and language editing, literature checks, code implementation and testing, and experimental analysis. Qwen and Tripo image-generation tools assisted with conceptual diagrams. Generative models used for candidate assets are described in the experimental setup.

{\small
\bibliographystyle{ieeenat_fullname}
\bibliography{references}
}
\clearpage
\appendix
\section*{Supplementary Material}
\addcontentsline{toc}{section}{Supplementary Material}
\label{sec:supplementary-material}
\section{Reading the evidence}
This supplement provides the experimental detail and visual evidence supporting the main findings. We first describe the evaluation populations and verification procedure, then external reconstruction references, alignment sensitivity, and reference controls. Subsequent sections report score aggregation, strong selection baselines, registration details, and the complete qualitative comparisons. Replica and ScanNet provide dense indoor RGB-D observations~\cite{straub2019the,dai2017scannet}; the ShapeR evaluation cohort and Digital Twin Catalog (DTC) provide additional reconstruction methods and capture settings~\cite{siddiqui2026shaper,dong2025digital}.

For decision rules and the scene-disjoint acceptance test, see \secref{sec:verification-update}. External camera calibration and cross-method choice appear in \secref{sec:mechanism-1101}; reference interventions and screening controls appear in \secref{sec:reference-controls}. Score aggregation and failure handling are detailed in \secref{sec:saggregation} and \secref{sec:spenalty}. The complete per-scene and image comparisons begin in \secref{sec:svisual-selection}.

\begin{table*}[t]
\centering
\caption{\textbf{Evaluation populations.} Counts across rows are not additive. Dense asset-choice populations are nested in the fixed-state study. External output availability and completed surface evaluation have separate denominators.}
\label{tab:scohorts}
\small
\begin{tabular}{p{0.20\textwidth}p{0.28\textwidth}p{0.25\textwidth}p{0.17\textwidth}}
\toprule
Study & Independent identity/group structure & Observation count & Use \\
\midrule
Original capture protocol & 137 instances; 10 scenes & 668 instance--view pairs & Fixed-state measurement \\
Selection-eligible & 134 instances; same 10 scenes & 268 calibration + 396 held-out & Availability-aware choice \\
Calibration-paired selection & 90 instances: 30 Replica + 60 ScanNet & Nested held-out views & Shared-candidate ranking \\
ShapeR-native inventory & 178 objects; 7 scenes in source cohort & 178 requests per route & Output coverage \\
DTC-native inventory & 100 objects; 2 captures each & 200 records per route & Output coverage \\
Multiview surface evaluation & 125 test objects; 5 scenes & 250 recorded-frame + 122 registered candidates & Surface-based choice \\
DTC catalog-associated & 16 dev / 54 test identities & Active capture only & Alignment and screening \\
\bottomrule
\end{tabular}
\end{table*}

\begin{figure*}[t]
\centering
\includegraphics[width=0.92\textwidth]{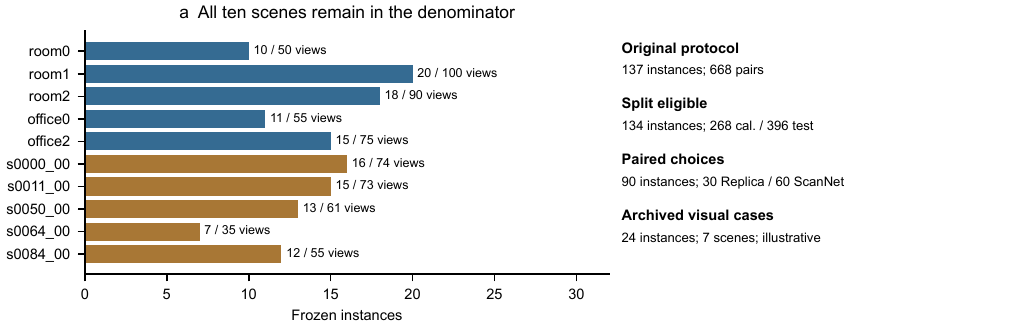}
\caption{\textbf{Scene coverage and nested evaluation populations.} All ten scenes and 668 views remain in the frozen inventory. Selection eligibility, paired ranking, and archived visual examples have distinct denominators. The 24 displayed instances cover seven scenes; quantitative results cover all five scenes in each domain.}
\label{fig:scoverage}
\end{figure*}

\begin{table*}[t]
\centering
\caption{\textbf{Evidence scope across the scientific questions.} Each row separates the dimension tested, its measured endpoint, and the strongest conclusion supported by the current records.}
\label{tab:evidence_scope}
\small
\begin{tabular}{p{0.13\textwidth}p{0.23\textwidth}p{0.25\textwidth}p{0.27\textwidth}}
\toprule
Question & Contrast and denominator & Evidence & Supported conclusion \\
\midrule
Does capture correspondence fail? & Tripo/Hunyuan replacement; 137 selected instances, 10 scenes & MaskIoU, visible coverage, overlap-depth; two capture domains & Both generators show measurable representation-replacement gaps in Replica and ScanNet. \\
Can registration explain the gap? & Anchor, calibration-transfer, and free-fit sensitivity on shared valid rows & Chamfer and F-score under identical geometry keys & Calibration transfer reduces partial-reference distance; residual error remains and anchored error is not pure shape error. \\
Does measurement alter asset choice? & Geometry and observation-profile choices on the same 90 paired instances & Calibration-only predictions against held-out-view IoU & The profile gets 78 choices correct versus 65 for geometry; the post-hoc IoU-only rule gets 85. \\
Does the finding extend across domains and generators? & Replica/ScanNet and Tripo/Hunyuan & Paired fixed-state measurements and per-domain choice outcomes & Replacement gaps appear in both domains; backend ordering reverses by domain. \\
Does it extend across reconstruction methods? & 122 shared multiview objects; 54 DTC test captures & Surface-based choice, method-specific registration, retained-pose outcomes & Observations improve choice across three methods; reference controls separately examine absolute screening. \\
Are category effects established? & Frozen instance identities and available source metadata & Object-name tokens in IDs; no verified taxonomy-to-instance join & The visual examples span varied named objects, but category-level rates and balanced generalization are not established. \\
\bottomrule
\end{tabular}
\end{table*}

\paragraph{Missing observations and outputs.}
Mesh availability, valid placement, reference support, and successful measurement are reported separately. Every requested comparison retains either its score or its reason for exclusion.

\section{Alignment-aware verification and acceptance}
\label{sec:verification-update}
The verification extension distinguishes choosing the better candidate from establishing that a candidate meets a capture requirement. The former is a relative ranking problem; the latter permits all candidates to be excluded. Both use calibration observations for decisions and open final-test observations only after the decision artifact is frozen.

\subsection{Observed surfaces, free space, and unknown support}
Let $M_v$ contain reliable object pixels with valid observed camera-depth $D_v$, and let $N_v$ contain valid-depth pixels reliably outside the object. Boundary pixels are excluded by a one-pixel band. For rendered candidate depth $\widehat D_{mv}(u;T)$, we measure
\begin{align}
 e_s(u)&=\min\!\left(1,\frac{|\widehat D_{mv}(u;T)-D_v(u)|}{\epsilon}\right),\quad u\in M_v,\\
 e_f(u)&=\min\!\left(1,\frac{[D_v(u)-\widehat D_{mv}(u;T)-\epsilon]_+}{\epsilon}\right),\quad u\in N_v.
\end{align}
A missing candidate intersection on an object ray contributes one to $e_s$; a missing intersection on a nonobject ray contributes zero to $e_f$. A candidate behind an observed nonobject occluder has no free-space penalty. The two terms are summed separately and normalized by $|M_v|$; their sum defines $L_v$. The replay uses $\epsilon=0.01$\,m and averages view losses. Support counts accompany each measurement.

Sparse captured points require a different observation interface. A measured ray comprises its world-space optical origin, unit direction, measured range, and object/nonobject/unknown label. First mesh intersections are compared with those ranges directly, preserving metre units under candidate transforms. Unmeasured image pixels remain unknown. Sparse object rays alone provide a surface term; the free-space term requires independently associated nonobject rays. A missing nonobject observation is recorded as unavailable. Dense-mask IoU is evaluated only with dense support annotations.

\subsection{Alignment envelopes and decision actions}
Four yaw starts and four spatial-block resamples produce up to 16 calibration fits per candidate. The retained set satisfies
\begin{equation}
 \mathcal H_m=\{T:L_{\calV}(m,T)\leq\min_{T'} L_{\calV}(m,T')+\delta\},\quad \delta=0.05.
\end{equation}
Its minimum and maximum losses define $[\ell_m,u_m]$. The relative decision selects $m$ when $u_m+\gamma<\ell_n$ for every rival $n$, using $\gamma=0$ in the recorded runs. Thus every retained pair of transforms preserves the ranking. This is a property of the finite evaluated set; held-out observations separately measure its transfer.

Absolute requirements constrain individual evidence components. A candidate passes when every retained hypothesis satisfies every requirement, fails when every retained hypothesis has a measured violation, and remains ambiguous otherwise. An unavailable required component leaves feasibility unresolved. The decision accepts a uniquely feasible candidate when all rivals fail, or a separated winner among feasible candidates. It excludes all candidates when each fails. Remaining ambiguity produces inspection while query budget remains and abstention when it is exhausted. Minimax fallback transforms supply transfer diagnostics separately from actual candidate decisions.

\subsection{Controlled defects establish measurement behavior}
\begin{figure*}[t]
\centering
\includegraphics[width=\textwidth]{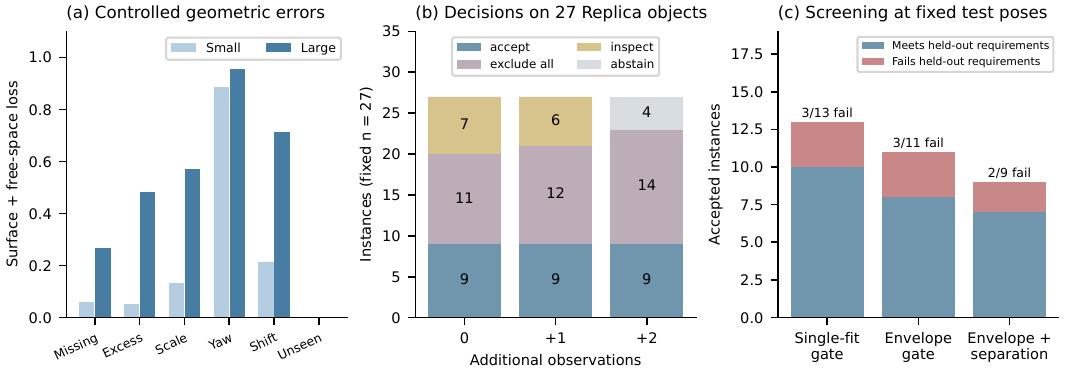}
\caption{\textbf{Controlled errors and asset-screening decisions.} (a) Two magnitudes of six edits to a known synthetic object. (b) Cumulative actions on the same 27 Replica instances; early stopping preserves the population at each observation budget. (c) Initial-budget acceptances and held-out violations with the same test pose for each candidate. Bars show descriptive counts with different accepted subsets.}
\label{fig:verification-update}
\end{figure*}

A known box and fixed cameras support six deterministic edit families: missing surface, added component, local scale, orientation, translation, and unseen-side completion. Thirteen predeclared checks pass. Enlarging a visible hole increases surface loss from 0.058 to 0.268; enlarging an added component increases free-space loss from 0.054 to 0.483. An unseen rear addition has zero measured effect. Calibration-only translation removes a 0.16\,m placement shift, reducing its loss from 0.713 to zero, while the missing-surface loss remains 0.268. These interventions distinguish correctable placement from persistent representation errors (\figref{fig:verification-update}a).

\subsection{Scene-disjoint Replica acceptance}
The development set contains three room2 instances. The recorded surface and free-space ceilings are 0.98795 and 0.04752, obtained component-wise from the better worst-case development candidate. The thresholds use calibration measurements; no final-test observation sets either value. Four other scenes supply 27 test instances. These ceilings define the reported development-relative acceptance rule, while the original asset-choice study retains its own ranking outcome and denominator.

At budget zero, nine candidates are accepted, 11 instances exclude all candidates, and seven require inspection. After two additional observations, the cumulative counts are nine accepted, 14 excluded, and four abstained (\figref{fig:verification-update}b). The accepted set remains unchanged. Two of the nine accepted candidates fail the same requirements on held-out views; one excluded instance has a passing alternative. Eight accepted choices match the archived-anchor ranking oracle. On those same nine identities, IoU selection matches nine oracle choices and geometry matches seven. Ranking correctness and requirement satisfaction therefore answer distinct questions.

The common-pose comparison evaluates every candidate at its calibration-best ray transform. Single-fit screening accepts 13 candidates with three held-out violations; alignment-set screening with separation accepts nine with two violations (\figref{fig:verification-update}c). On those same nine identities, single-fit screening also yields two violations and mean regret 0.03098. Forced IoU choice yields two violations and regret 0.02170. Comparing the same accepted identities therefore distinguishes screening improvements from reduced coverage.

Twelve of 54 candidate alignment sets contain a single distinct transform. Set-based screening changes an action or candidate relative to single-fit screening on seven instances: it defers one noncompliant and three compliant acceptances, and converts three correct exclusions into unresolved decisions. These cases identify the sensitivity--coverage tradeoff introduced by retaining alternative alignments.

\section{External reconstruction: reference construction and alignment}
\label{sec:mechanism-1101}
We evaluate ShapeR, Amodal3R, and SceneGen on the ShapeR multiview cohort~\cite{siddiqui2026shaper,wu2025amodal3r,meng2025scenegen}, and ShapeR/EFM3D on DTC~\cite{dong2025digital,straub2024efm3d}. Measured surface points, calibrated optical rays, and method-specific coordinate mappings connect outputs to capture observations. Calibration observations determine registration and candidate choice; held-out observations evaluate the resulting surfaces.

\subsection{Object association and capture availability}
DTC object support is associated with the posed catalog mesh using a 2\,cm point-to-surface radius. These labels are catalog-derived. EFM3D's fused mesh is already in world coordinates. For the evaluated ShapeR outputs, let $T$ map the catalog object frame into the world frame. Their recorded output convention applies $T^{-1}$ during normalization; applying $T$ to undo this operation and then $T$ to place the object gives the mapping $T\,T$.

Catalog references are available for 165 of 179 candidate capture records. Requiring at least 12 object rays retains 16 of 32 development and 54 of 147 test captures. Development omits 14 captures with insufficient support and two without a reference association; the corresponding test counts are 81 and 12. Every retained capture is active. The results therefore describe active capture, with no paired active/passive comparison under this reference construction.

\paragraph{Camera calibration.}
DTC rays use the recording's Fisheye624 model and camera-to-device extrinsics composed with the device trajectory. Camera serials match point observations to the calibrated optical stream. Valid views require timestamp disagreement at most 1\,ms, 95th-percentile reprojection error at most 2\,px, lateral residual at most 2\,cm, and at least 12 object and 12 background rays. These checks test optical consistency separately from catalog-based object association.

\begin{table}[t]
\centering
\caption{\textbf{DTC alignment sensitivity.} All 54 test captures are active. Multi counts candidates with multiple distinct retained transforms; Range uses summed surface/free-space loss. Sat. counts calibration-best poses with held-out surface loss exactly one.}
\label{tab:dtc-1101}
\small\begin{tabular}{lrrrr}
\toprule
Method & $N$ & Multi & Range$>$.05 & Sat. \\
\midrule
EFM3D & 54 & 11 & 7 & 18 \\
ShapeR & 54 & 27 & 21 & 12 \\
\bottomrule
\end{tabular}

\end{table}

Across 108 candidate alignment sets, 38 retain multiple poses and 28 have held-out loss ranges above 0.05 (\tabref{tab:dtc-1101}). Thus calibration-compatible placements can lead to different unseen surfaces. The screening comparison below measures whether retaining this variation improves decisions.

\subsection{Multiview surface references and SceneGen registration}
Measured object points and their optical centers define reference rays for the ShapeR evaluation cohort. These rays constrain object surfaces; background rays and dense silhouette masks are unavailable. ShapeR and Amodal3R are evaluated on 53 development and 125 test objects.

SceneGen predicts normalized scene geometry after moving the query object to the origin and scaling the scene extent to 1.96. Calibration registration initializes translation and uniform scale from measured object points and mesh bounds, followed by the same multistart fitting procedure. Registration is available for 52 of 53 development and 122 of 125 test objects; four objects lack the required calibration association. Registration and candidate decisions are fixed before test observations are used. This comparison evaluates the registered object surface, with SceneGen's original scene layout outside the evaluated quantity.

\begin{table}[t]
\centering
\caption{\textbf{Surface evidence across three reconstruction methods.} Test objects span five scenes. ShapeR/Amodal3R use recorded coordinate frames; SceneGen uses calibration registration. Range is the retained-pose held-out surface-loss range. Sat. counts calibration-best poses with held-out loss exactly one.}
\label{tab:b6-1101}
\small\begin{tabular}{lrrrr}
\toprule
Method & $N$ & Multi & Range$>$.05 & Sat. \\
\midrule
ShapeR & 125 & 124 & 1 & 91 \\
Amodal3R & 125 & 124 & 3 & 2 \\
SceneGen (cal.) & 122 & 121 & 10 & 0 \\
\bottomrule
\end{tabular}

\end{table}

SceneGen has 121 candidates with multiple retained poses and ten held-out loss ranges above 0.05. The corresponding counts are 124/3 for Amodal3R and 124/1 for ShapeR (\tabref{tab:b6-1101}). ShapeR has 91 saturated surface losses and median missing-object-ray fraction one. We therefore retain missing intersections separately from depth errors on intersecting rays: the same bounded loss can arise from either source.

\subsection{Asset selection across reconstruction methods}
The three routes overlap on 122 test objects. We perform a post-hoc comparison of geometry choice, calibration-best surface choice, and upper-envelope surface choice. All use the same observations and stored fitting budget. Geometry first selects its own observed-to-mesh-best pose and candidate. The common-deployment control then evaluates that chosen candidate at its calibration-best surface pose, as used by the other rules. The oracle compares these same deployment poses. This separation prevents a different deployment pose from being credited as a ranking gain.

\begin{table}[t]
\centering
\caption{\textbf{Post-hoc three-route choice at common deployment poses.} Correctness and regret use the candidate with minimum held-out surface loss. All rules choose among the same three candidates; no absolute acceptance requirement is imposed.}
\label{tab:b6-choice-1101}
\small\begin{tabular}{lrrr}
\toprule
Rule & Correct & Loss $\downarrow$ & Regret $\downarrow$ \\
\midrule
Fixed route (dev.) & 63 & 0.9379 & 0.01653 \\
Geometry selection & 91 & 0.9263 & 0.00484 \\
Single-fit surface & 119 & 0.9214 & 0.00002 \\
Worst-case surface & 93 & 0.9238 & 0.00241 \\
\bottomrule
\end{tabular}

\end{table}

Single-surface choice selects the held-out-best candidate for 119/122 objects; geometry selects it for 91/122, and upper-envelope choice for 93/122. A fixed-route baseline selects SceneGen from the lowest mean development held-out loss on 52 shared development objects, then uses that route for every test object: it is correct on 63/122. Thus the observation rule improves on both geometry and a development-selected generator prior. Geometry's own-pose mean surface loss is 0.9427, versus 0.9263 at common deployment. Single and envelope means are 0.9214 and 0.9238. The upper envelope does not improve on the strong single-surface rule. All means remain near the bounded-loss ceiling, so relative choice and absolute asset usability remain distinct outcomes. SceneGen's initialization uses calibration points and is reported separately from native-frame initialization.

\begin{figure*}[t]
\centering
\includegraphics[width=\textwidth]{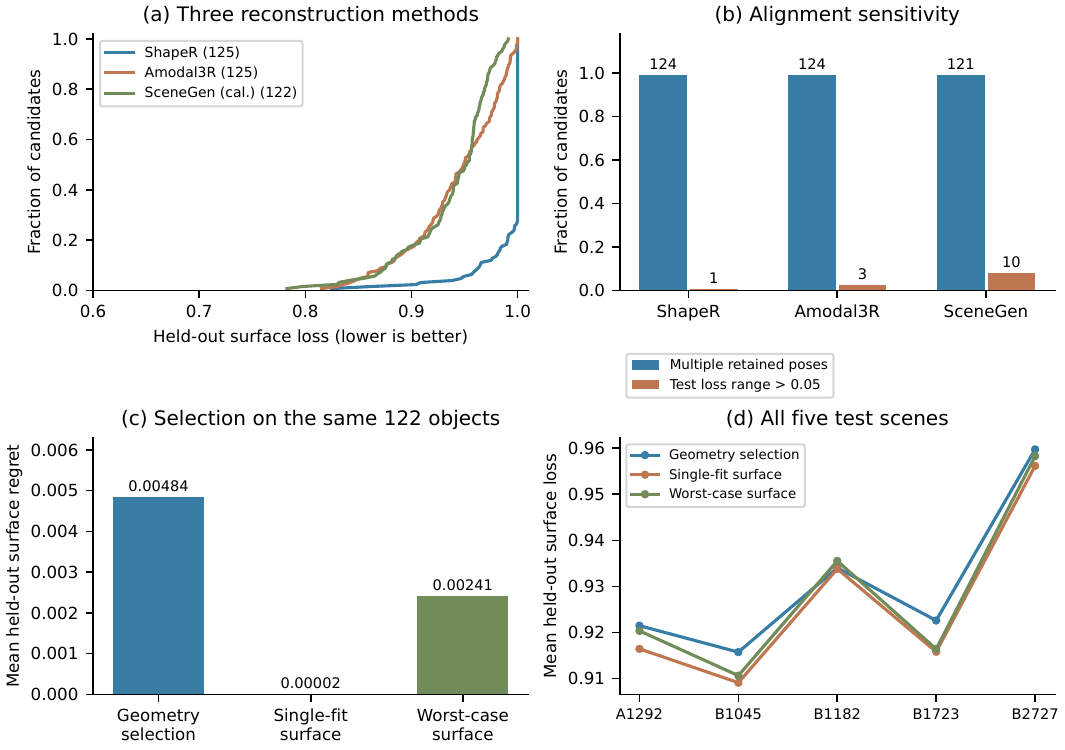}
\caption{\textbf{Surface-based selection across reconstruction methods.} (a) Held-out surface-loss distributions, including saturation. (b) Multiple compatible poses are frequent, while larger unseen loss ranges are less common; denominators appear in \tabref{tab:b6-1101}. (c) All selectors use the same 122 objects and candidate-specific test poses. (d) Per-scene results include reversals in selector ordering. This post-hoc comparison evaluates relative surface quality, separately from absolute acceptance and scene-layout accuracy.}
\label{fig:cross-route-1101}
\end{figure*}

\subsection{Fitting-budget and evidence sensitivity}
Sensitivity analysis uses the same 16 DTC development objects. One, four, or 16 yaw starts, each with four spatial resamples, produce four, 16, or 64 fitting proposals per candidate. Every configuration additionally evaluates the initial transform and six 2\,cm translation probes. A point-count control reduces registration samples from 4,000 to 1,000 while preserving the reference rays. A tolerance control increases the depth tolerance from 1\,cm to 5\,cm. These changes test registration and measurement sensitivity separately from candidate-screening accuracy.

\begin{table}[t]
\centering
\caption{\textbf{Sensitivity to fitting budget and depth tolerance.} Multi counts candidates with multiple distinct retained transforms. Time is the median per-object measurement time including both methods. These development comparisons evaluate measurements rather than screening accuracy; cache effects are included in the timing.}
\label{tab:sensitivity-1101}
\small\begin{tabular}{lrrrr}
\toprule
Configuration & $N$ & Fits & Multi & s \\
\midrule
Default & 16 & 16 & 12 & 1.73 \\
1 yaw, 4 resamples & 16 & 4 & 12 & 0.65 \\
16 yaw, 4 resamples & 16 & 64 & 14 & 3.91 \\
5 cm evidence tolerance & 16 & 16 & 16 & 1.42 \\
1,000 fitting points & 16 & 16 & 12 & 0.85 \\
\bottomrule
\end{tabular}

\end{table}

\section{Reference controls and absolute quality requirements}
\label{sec:reference-controls}
\subsection{Visible errors and unobserved geometry}
First-visible-surface association requires points to lie within 2\,cm of the catalog surface and their range to agree with its first ray intersection within 2\,cm. Occluded points become unknown. A local-defect control moves vertices within 5\,mm of the first object-ray hit by 2\,cm along that ray. Moving this neighborhood prevents rays from falling onto nearby stacked triangles after an isolated triangle is displaced. A separate 5\,cm translation tests placement error; an unobserved edit tests invariance beyond the measured surfaces.

Quality thresholds are the 95th percentile of development reference losses plus fixed component slack (0.02 surface, 0.05 free space), giving bounds 0.6362 and 0.1548 under first-visible-surface association. Candidate defects and test observations do not determine these thresholds.

\begin{table}[t]
\centering
\caption{\textbf{Development controls with first-visible-surface support.} Expected means reference/edit acceptance or visible-defect detection. Unchanged tests equality to the corresponding reference, independently of acceptance.}
\label{tab:controls-1101}
\small\begin{tabular}{lrr}
\toprule
Control & Expected & Unchanged \\
\midrule
Catalog reference & 15/16 & -- \\
5 cm placement & 16/16 & -- \\
2 cm local surface & 16/16 & -- \\
Unobserved edit & 15/16 & 16/16 \\
\bottomrule
\end{tabular}

\end{table}

All 16 placement errors and all 16 local surface defects exceed the requirements. All 16 unobserved edits preserve both evidence components exactly. Fifteen references satisfy the requirements; the same fifteen unobserved edits satisfy them (\tabref{tab:controls-1101}). Gargoyle accounts for the two unmet checks among 64 controls: its reference surface loss is 0.6779, above the 0.6362 bound, and its identical unobserved edit shares that outcome. The complete reference-validity audit therefore remains unmet, while the visibility invariance and both visible-error controls hold on every tested identity.

\begin{figure*}[t]
\centering
\includegraphics[width=\textwidth]{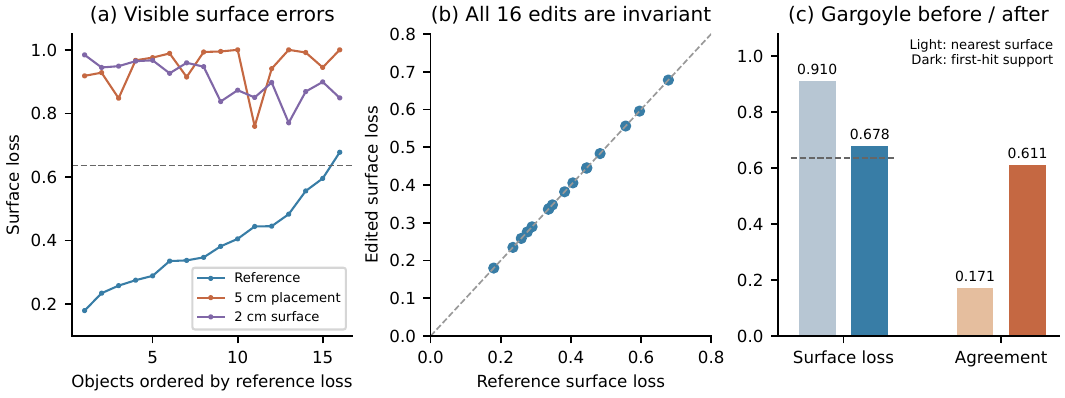}
\caption{\textbf{Detection, visibility invariance, and reference association.} (a) The two visible interventions increase surface loss beyond the development requirement; the reference ordering retains every identity. (b) Unobserved edits leave all 16 reference measurements unchanged, including the reference above the quality bound. (c) First-hit filtering improves Gargoyle's surface agreement, while its remaining discrepancy still exceeds the bound. Dashed lines in (a,c) indicate the surface requirement.}
\label{fig:reference-controls-1101}
\end{figure*}

Gargoyle illustrates why reference association is part of the measurement. First-hit filtering moves 186 of 258 former object rays to unknown. Surface agreement rises from 0.171 to 0.611, overlap depth error falls from 4.74\,cm to 0.84\,cm, and surface loss falls from 0.9101 to 0.6779. The free-space violation numerator is unchanged; normalizing it by 72 rather than 258 object rays raises its component from 0.0349 to 0.1250. This increase reflects the changed denominator. A reference quantile plus fixed slack also permits a tail reference to remain above the requirement; we preserve that identity and the original derivation policy.

\subsection{Single-fit and alignment-set screening}
The DTC screening comparison uses nearest-surface object association and its development bounds (0.8195 surface, 0.1166 free space). It comprises 16 development and 54 test captures. The first-visible-surface controls above use a separate reference definition and corresponding bounds; they are evaluated independently of these candidate decisions.

\begin{table}[t]
\centering
\caption{\textbf{DTC screening with nearest-surface reference association.} All rules evaluate the same 54 active test captures. Out counts accepted assets; Bad counts those violating held-out requirements. These decisions use the reference and thresholds specified in this subsection.}
\label{tab:acceptance-1101}
\small\begin{tabular}{lrrrr}
\toprule
Rule & Out & Bad & Excl. & Defer \\
\midrule
Geometry & 4 & 3 & 0 & 50 \\
Single-fit & 10 & 8 & 44 & 0 \\
Alignment set & 10 & 8 & 44 & 0 \\
Set + separation & 10 & 8 & 44 & 0 \\
\bottomrule
\end{tabular}

\end{table}

Geometry accepts four assets with three held-out violations. Single-fit, set-based, and set-based separated screening each accept ten with eight violations and exclude every candidate for 44 captures (\tabref{tab:acceptance-1101}). Single-fit and set-based separated screening choose identical actions and candidates on every test capture. Forty-three exclusions remove only noncompliant candidates; one withholds a compliant candidate. All registered fixed margins yield six development acceptances and six violations, so none meets the target 10\% risk. The alignment set supplies no screening improvement in this comparison. Reliability under the first-visible-surface reference remains a separate, unevaluated candidate-screening condition.

\begin{table*}[t]
\centering
\caption{\textbf{Support and surface components of the original replacement gap.} Conditional instance-macro coverage loss and overlap-depth increase. Depth requires valid overlap; ScanNet/Hunyuan has 62 depth-valid instances and 63 silhouette instances. Intervals use the archived hierarchical procedure.}
\label{tab:sdiagnostics}
\small\begin{tabular}{llrll}
\toprule
Domain & Backend & $n$ & Coverage loss (95\% CI) & Depth increase / m (95\% CI) \\
\midrule
Replica & Tripo & 74 & 0.322 [0.258, 0.398] & 0.171 [0.108, 0.251] \\
Replica & Hunyuan & 30 & 0.254 [0.163, 0.337] & 0.177 [0.087, 0.294] \\
ScanNet & Tripo & 63 & 0.304 [0.213, 0.394] & 0.122 [0.067, 0.182] \\
ScanNet & Hunyuan & 63 & 0.396 [0.293, 0.483] & 0.283 [0.242, 0.327] \\
\bottomrule
\end{tabular}

\end{table*}

\section{Evaluation setup and aggregation}
\label{sec:saggregation}
\subsection{Object identity and coordinate conventions}
An observation row is keyed by dataset, scene/capture ID, instance ID, candidate ID, and view ID. The candidate record identifies the asset checksum and transform. The reference record specifies the observation, support annotation, depth, camera calibration, point-cloud checksum, world frame, and metre units. Each endpoint identifies its placement convention, including native output, archived anchor, calibration transfer, or free per-view fitting. Changes to assets, references, or transforms retain a distinct endpoint identity.

Connecting a new method requires resolving object identity and native coordinates before rendering. Object-conditioned reconstruction may supply an isolated mesh, whereas a fused scene prediction requires an object-extraction rule. Each native frame is mapped explicitly to the capture frame. Separate asset and transform checksums preserve the geometry and placement used by the evaluator; identity transforms use a documented coordinate convention.

\subsection{View-disjoint selection}
The selection split assigns two calibration views to each eligible instance. It retains the remaining views for scoring: 134 instances yield 268 calibration and 396 held-out views. Recorded generation inputs are reserved to calibration when their identities are recoverable; stable ordering determines the remaining membership. Three ScanNet instances, totaling four original views, cannot satisfy the required split and are excluded from selection but remain in the original 137-instance measurement protocol. No generated score determines this eligibility.

For each candidate, the geometry rule averages its calibration-transfer Chamfer values on calibration observations and minimizes the result. The observation rule maximizes the archived anchored three-component calibration profile, not IoU alone. The choice is then frozen. The retrospective oracle maximizes the held-out mean. Predictor ties use lexical candidate ordering; oracle ties count any equally best chosen candidate as correct. Random choice is the analytical uniform expectation on the same candidates.

The principal ranking cohort contains 90 instances, with both candidates eligible at calibration for both selectors. Its 30 Replica and 60 ScanNet instances compare choices between the same available alternatives. Held-out visibility is evaluated after the choice and contributes to its outcome. The all-eligible 134-instance analysis additionally captures the effect of candidate availability at calibration.

\subsection{Aggregation and missingness}
For IoU, each instance receives the mean over its specified views, and the domain aggregate is the mean over instances. Frame-rich objects therefore do not dominate. Coverage first sums intersecting and reference pixels within an instance, then divides those sums. Overlap-depth error pools valid overlapping depth pixels within an instance; an instance with no valid overlap has no finite depth error. The domain depth mean consequently has a declared valid-instance denominator.

The original fixed-state full-inventory score retains missing renders with zero overlap. Its conditional summary uses a separately identified visible-asset cohort. For Hunyuan/Replica, that cohort has 30 instances rather than 74; the 44 missing meshes are not deleted from the inventory. \tabref{tab:savailability} reports both asset and view failure counts. The one ScanNet/Hunyuan instance without valid depth overlap remains in the silhouette count of 63, but not in the depth count of 62.

\begin{table*}[t]
\centering
\caption{\textbf{Availability under the original frozen protocol.} Instances and missing meshes refer to the requested inventory. Non-OK views include missing or invisible endpoint renders; a non-OK render does not by itself establish a missing mesh.}
\label{tab:savailability}
\small\begin{tabular}{llrrrr}
\toprule
Domain & Backend & Instances & Views & Missing GLB & Non-OK views \\
\midrule
Replica & Tripo & 74 & 370 & 0 & 10 \\
Replica & Hunyuan & 74 & 370 & 44 & 220 \\
ScanNet & Tripo & 63 & 298 & 0 & 0 \\
ScanNet & Hunyuan & 63 & 298 & 0 & 6 \\
\bottomrule
\end{tabular}

\end{table*}

\section{Failure-policy sensitivity of asset choice}
\label{sec:spenalty}
The archived selection run defines view utility as $u=s$ for a successful MaskIoU observation and $u=-1$ otherwise. A candidate's held-out utility is the mean of these view utilities. This is not a bounded IoU: for example, one successful view scoring 0.2 and one missing view have utility $-0.4$. Assigning zero to each missing view instead gives mean IoU 0.1. Clipping the already averaged utility to zero would give a third, incorrect quantity for that rescore.

We therefore report the original utility and a post-hoc zero-failure sensitivity. The latter reuses exactly the same stored calibration decisions, substitutes zero at each non-OK held-out \emph{view}, then repeats view-to-instance aggregation and computes oracle regret. It does not rerank candidates at calibration, fit a new threshold, or retune a predictor. Four paired instances contain a negative candidate aggregate under the original utility; the distinction changes the magnitude of geometry regret.

\begin{table*}[t]
\centering
\caption{\textbf{Original failure-penalized utility on the paired cohort.} This preserves the archived $-1$-per-failure outcome. The main paper instead shows the unchanged choices rescored with zero missing-view IoU. Accuracy is unchanged in this cohort, whereas the regret units and magnitudes differ.}
\label{tab:spenalty}
\small\begin{tabular}{lrlrrr}
\toprule
Domain & $n$ & Selector & Correct & Acc.\,(\%) $\uparrow$ & Utility regret $\downarrow$ \\
\midrule
Replica & 30 & Cal.-transfer CD & 19/30 & 63.3 & 0.0608 \\
Replica & 30 & Cal. profile & 26/30 & 86.7 & 0.0062 \\
ScanNet & 60 & Cal.-transfer CD & 46/60 & 76.7 & 0.0210 \\
ScanNet & 60 & Cal. profile & 52/60 & 86.7 & 0.0150 \\
Pooled & 90 & Cal.-transfer CD & 65/90 & 72.2 & 0.0343 \\
Pooled & 90 & Cal. profile & 78/90 & 86.7 & 0.0121 \\
\bottomrule
\end{tabular}

\end{table*}

\begin{table}[t]
\centering
\caption{\textbf{All-eligible choice, original utility.} The 134-instance cohort includes candidates unavailable at calibration, unlike the paired ranking cohort.}
\small\begin{tabular}{lrrr}
\toprule
Selector & $n$ & Acc.\,(\%) & Utility regret \\
\midrule
Cal.-transfer CD & 134 & 81.3 & 0.0230 \\
Cal. profile & 134 & 91.0 & 0.0081 \\
\bottomrule
\end{tabular}

\end{table}

\begin{table}[t]
\centering
\caption{\textbf{All-eligible choice, zero-failure sensitivity.} Only the held-out failure outcome is changed, not the recorded choices.}
\small\begin{tabular}{lrrr}
\toprule
Selector & $n$ & Acc.\,(\%) & IoU regret \\
\midrule
Cal.-transfer CD & 134 & 81.3 & 0.0180 \\
Cal. profile & 134 & 91.0 & 0.0081 \\
\bottomrule
\end{tabular}

\end{table}

\paragraph{Asset-choice policy.}
The predictor in the original ranking study selects one candidate and retains every selected asset. Its held-out suitability threshold of 0.5 supplies outcome diagnostics. That study evaluates which available candidate best explains unseen observations. The separate acceptance experiments in \secref{sec:verification-update} freeze quality requirements on development groups before evaluating held-out risk.

\subsection{Archived selector and post-hoc component controls}
The archived predictor is the equal-weight mean of per-view IoU, coverage, and clipped depth quality, with a 0.05\,m depth cap. Its calibration average includes $-1$ for a failed profile view. Reconstructing that formula and all archived calibration candidate means reproduces the recorded decisions. It is not an IoU-only selector.

The component controls use calibration rows only, with the same paired candidates and held-out zero-failure outcome. Failed IoU/coverage calibration views receive $-1$; failed depth receives positive infinity, so a candidate with a failed depth view cannot beat one with finite mean depth. Lexical candidate ID resolves exact feature ties. These are additional post-hoc rules, without weight tuning on held-out outcomes. They are not retrospective replacements for the archived profile. Depth clipping can saturate, and a component that diagnoses an error need not improve a scalar decision rule.

\begin{table*}[t]
\centering
\caption{\textbf{Same-denominator selector component analysis.} Calibration features select; held-out IoU supplies correctness and regret. Missing held-out views contribute zero. The three component-only rules are post-hoc; the profile and transfer-CD choices are archived. Counts are instances, not independent model reruns.}
\label{tab:scomponents}
\small\begin{tabular}{llrrr}
\toprule
Domain & Calibration selector & $n$ & Correct & IoU regret \\
\midrule
Replica & Transfer CD & 30 & 19/30 & 0.0386 \\
Replica & Three-component profile & 30 & 26/30 & 0.0062 \\
Replica & IoU only (post-hoc) & 30 & 27/30 & 0.0003 \\
Replica & Coverage only (post-hoc) & 30 & 28/30 & 0.0001 \\
Replica & Depth only (post-hoc) & 30 & 18/30 & 0.0357 \\
ScanNet & Transfer CD & 60 & 46/60 & 0.0210 \\
ScanNet & Three-component profile & 60 & 52/60 & 0.0150 \\
ScanNet & IoU only (post-hoc) & 60 & 58/60 & 0.0034 \\
ScanNet & Coverage only (post-hoc) & 60 & 45/60 & 0.0404 \\
ScanNet & Depth only (post-hoc) & 60 & 44/60 & 0.0435 \\
Pooled & Transfer CD & 90 & 65/90 & 0.0269 \\
Pooled & Three-component profile & 90 & 78/90 & 0.0121 \\
Pooled & IoU only (post-hoc) & 90 & 85/90 & 0.0024 \\
Pooled & Coverage only (post-hoc) & 90 & 73/90 & 0.0270 \\
Pooled & Depth only (post-hoc) & 90 & 62/90 & 0.0409 \\
\bottomrule
\end{tabular}

\end{table*}

\section{Geometry and placement details}
\begin{table*}[t]
\centering
\caption{\textbf{Outcome-stratified examples from the held-out choice study.} The display rule selects the largest absolute held-out candidate margin within each domain and available outcome stratum. Both directions of selector disagreement, common successes, and the common failure are retained. T/H columns average all held-out views with missing views scored zero. \figref{fig:smultiviewdecisions} expands every example into its calibration and held-out observations.}
\label{tab:sdecisioncases}
\small\begin{tabular}{llrllrr}
\toprule
Domain & Scene & ID & CD choice & Profile choice & Held-out T & Held-out H \\
\midrule
Replica & office2 & 1 & Tripo & Hunyuan & 0.227 & 0.701 \\
Replica & office0 & 1 & Hunyuan & Tripo & 0.243 & 0.357 \\
Replica & office2 & 16 & Hunyuan & Hunyuan & 0.130 & 0.741 \\
ScanNet & scene0064\_00 & 4 & Tripo & Hunyuan & 0.159 & 0.409 \\
ScanNet & scene0050\_00 & 125 & Tripo & Hunyuan & 0.420 & 0.197 \\
ScanNet & scene0050\_00 & 1 & Tripo & Tripo & 0.805 & 0.077 \\
ScanNet & scene0050\_00 & 2 & Tripo & Tripo & 0.170 & 0.270 \\
\bottomrule
\end{tabular}

\end{table*}

The six geometry files use identical candidate-view keys for Chamfer and F-score under anchor, calibration-transfer, and free per-view fitting. Each file has 1,328 rows. The common successful set across all six contains 1,108 rows; the remaining 220 lack valid candidate geometry. Distances are undefined for absent geometry and are not assigned zero, which would make an absent mesh look perfect. The main table averages valid views within each instance and then instances on the common valid set.

\begin{table*}[!t]
\centering
\caption{\textbf{Archived pose-feature incremental control.} The baseline uses CAST-orientation/HYBRID observation features. The bootstrap uses 5,000 scene--instance replicates, seed 0; its interval applies to augmented-minus-baseline regret. ScanNet exhibits a ceiling case with no measured gain.}
\label{tab:spose}
\small
\begin{tabular}{lrrrrrl}
\toprule
Domain & $n$ & Base acc.\,(\%) & Aug. acc.\,(\%) & Base regret & Aug. regret & Regret difference, 95\% CI \\
\midrule
Replica & 74 & 79.7 & 94.6 & 0.0272 & 0.0003 & $-0.0269$ [$-0.0702$, $-0.0021$] \\
ScanNet & 60 & 98.3 & 98.3 & 0.0017 & 0.0017 & No observed difference \\
\bottomrule
\end{tabular}
\end{table*}

\paragraph{Units and reference support.}
Visible reference points are obtained by back-projecting masked recorded depth into the world frame in metres. Candidate surfaces use 10,000 samples with seed 0. Chamfer is the mean of the two mean nearest-neighbor Euclidean distances, without squaring. The manuscript converts metres to centimetres only for display. Precision is the fraction of candidate samples within 1\,cm of the reference; recall is the corresponding fraction of reference points near the candidate. F-score is $2pr/(p+r)$, or zero when its denominator is zero.

The candidate samples span the full triangle surface, while the reference contains visible captured points. Candidate-to-reference distance therefore includes completed surfaces outside the observed support; reference-to-candidate distance evaluates how closely the mesh explains that support. ShapeR's complete-reference normalized geometry uses a different reference regime~\cite{siddiqui2026shaper}. Reporting these regimes separately preserves their geometric interpretation and numerical scale.

\paragraph{Three transform contracts.}
The archived anchor is retained without fitting against held-out observations. Calibration transfer fits a uniform similarity using calibration reference geometry and applies that single transform to the scoring views. Free fitting uses each target view's reference independently. The latter can inspect held-out geometry and is therefore only a sensitivity endpoint. In particular, it cannot become a feature in the calibration-only choice table. Iterative fitting can have local minima; its residual need not be smaller than calibration-transfer residual for every metric or cell.

The original anchor uses a scan-derived centre, uniform extent scaling, and upright PCA yaw without primary ICP. Its stored transform defines the tested endpoint. Calibration-transfer and free-fit comparisons characterize sensitivity to this initialization. Free fitting remains an iterative sensitivity estimate, with no guarantee of globally optimal alignment.

\begin{table*}[t]
\centering
\caption{\textbf{Transfer to held-out geometry.} Instance-macro CD in centimetres on held-out views only. These values differ from the main all-view geometry table because calibration views are not included here. Candidate/view keys are shared across the three modes.}
\small\begin{tabular}{llrrrr}
\toprule
Domain & Candidate & $n$ & Anchor CD & Transfer CD & Free CD \\
\midrule
Replica & Tripo & 74 & 24.26 & 19.19 & 18.27 \\
Replica & Hunyuan & 30 & 16.33 & 14.37 & 15.23 \\
ScanNet & Tripo & 60 & 16.74 & 13.72 & 12.37 \\
ScanNet & Hunyuan & 60 & 28.85 & 24.95 & 24.11 \\
\bottomrule
\end{tabular}

\end{table*}

\subsection{Separate scene-held-out pose-feature control}
\begin{table*}[!t]
\centering
\caption{\textbf{Availability of pretrained reconstruction outputs.} Counts are object--method pairs; DTC's 200 captures represent 100 identities. Completed surface evaluations use the subsets in \tabref{tab:scohorts}. LIRM has no evaluated output.}
\label{tab:snative}
\small\begin{tabular}{llrr}
\toprule
Cohort & Method & Requested & Available \\
\midrule
ShapeR cohort & ShapeR & 178 & 178 \\
ShapeR cohort & Amodal3R & 178 & 178 \\
ShapeR cohort & SceneGen & 178 & 174 \\
ShapeR cohort & LIRM & 178 & 0 \\
DTC & ShapeR & 200 & 200 \\
DTC & EFM3D & 200 & 199 \\
\bottomrule
\end{tabular}

\end{table*}

The archived pose-feature experiment is a small statistical analysis, distinct from the direct geometry-versus-profile rule. Its baseline features are CAST orientation-only and HYBRID multiview MaskIoU. The augmented model adds anchored MaskIoU, coverage, and overlap-depth L1. Each metric supplies a successful-calibration mean, oriented so larger is better, and a calibration availability fraction.

Both variants use ridge regression with fixed $\alpha=1$, leaving one scene out within a dataset. Feature imputation, centering, and scaling use training-fold statistics only. Candidate predictions rank held-out choices, with lexical candidate labels breaking exact ties. The outcome assigns zero to missing observations. This low-dimensional statistical control tests the additional predictive signal in correspondence features.

\tabref{tab:spose} shows a domain-dependent incremental signal over the tested pose-observation features. The direct geometry comparison and the pose-feature control have distinct baseline features and remain separate evidence blocks.

\section{Reconstruction methods and output availability}
\label{sec:snative}
\tabref{tab:snative} reports native output/status records. ShapeR uses capture-conditioned object evidence; Amodal3R reconstructs an occluded object from image evidence; SceneGen jointly generates scene objects and layout~\cite{siddiqui2026shaper,wu2025amodal3r,meng2025scenegen}. EFM3D reconstructs scene-level geometry from egocentric sequences~\cite{straub2024efm3d}. The visual inspection preserves these distinct input and output regimes as the starting point for common evaluation.

\paragraph{ShapeR-178.}
ShapeR and Amodal3R each provide 178 outputs. SceneGen provides 174 outputs and has four missing-output cases. Per-object evaluation additionally requires calibration observations and an object-specific registration, yielding the measured populations in \tabref{tab:scohorts}. LIRM is unavailable because its evaluated release does not provide the required runnable pretrained model.

\begin{figure}[!ht]
\centering
\includegraphics[width=\textwidth]{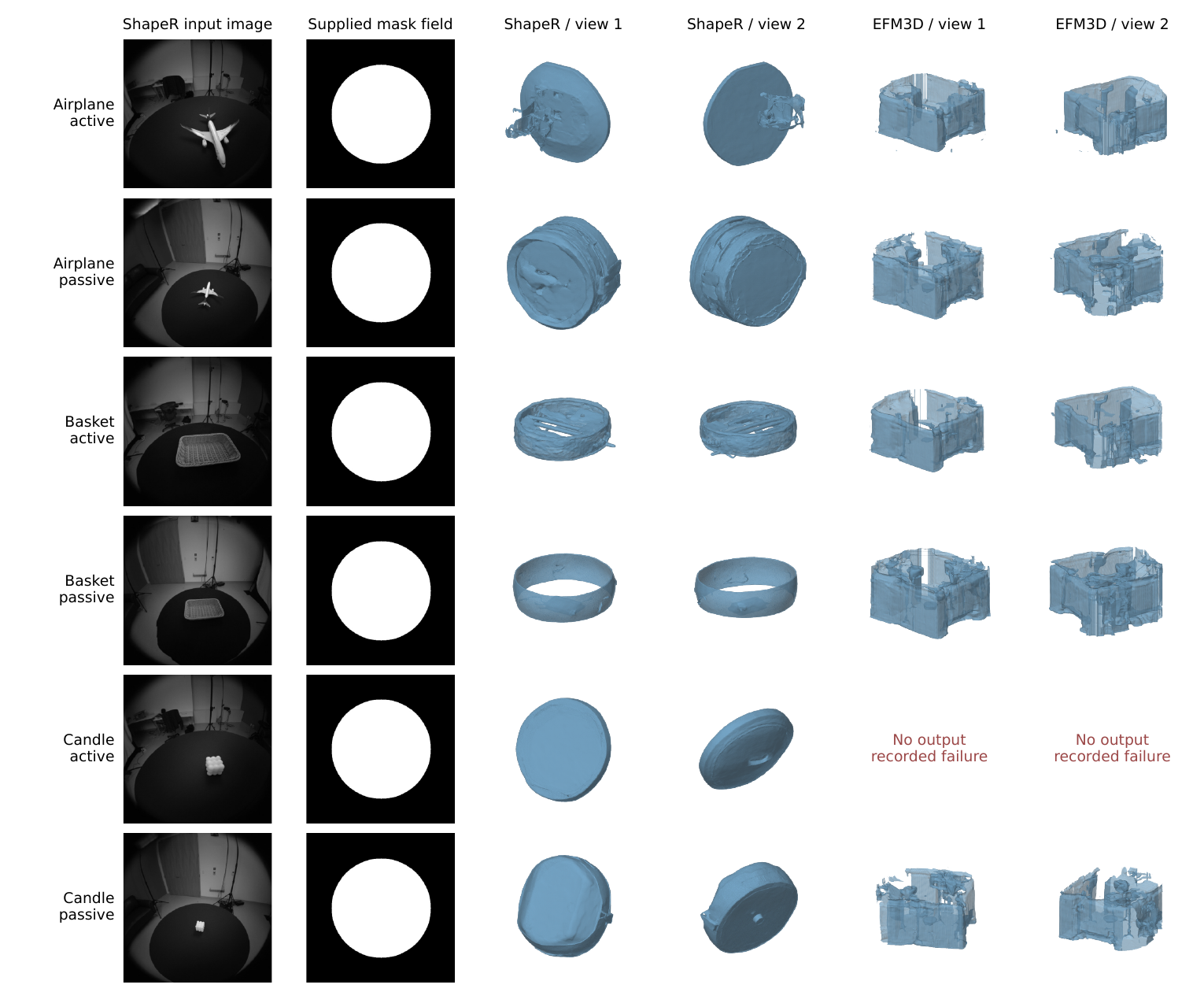}
\caption{\textbf{DTC input fields, active/passive native outputs, and a recorded failure.} Six rows connect three catalog objects in both capture modes to their archived ShapeR input images, supplied mask fields, and two native display views per method. The six input packets share the same mask encoding, intrinsics, and camera-to-world matrices, and each contains constant depth. This provenance audit identifies an input-adaptation issue to resolve before scoring capture correspondence. The input columns belong to ShapeR; they are not asserted to be EFM3D's inputs. The meshes use independently normalized native frames. EFM3D scene structure needs object extraction; its failed active-candle record remains explicit. The rows represent three identities and six captures.}
\label{fig:sdtcgallery}
\end{figure}

\paragraph{DTC.}
The DTC protocol supplies active and passive captures for 100 catalog object IDs. ShapeR has 200 successful records; EFM3D has 199 successful records and one active-capture failure associated with unreadable VRS pose data. The 178 ShapeR-cohort EFM3D records instead document an incompatible input route and remain outside the DTC denominator. We inspect six archived ShapeR input packets, covering three catalog objects in both capture modes (\figref{fig:sdtcgallery}). All six contain identical encoded mask fields, constant depth, identical intrinsics, and identical camera-to-world matrices. Their masks provide circular support across distinct objects. Common-quality comparison requires object-specific support and calibrated capture evidence. Repairing fields used by inference also requires rerunning the affected pretrained inference with those corrected inputs. The observed pattern applies to the six inspected packets.

\paragraph{Coordinates and object extraction.}
Output availability and calibrated evaluation have separate requirements. Object meshes require a local-to-capture transform, while scene-level surfaces additionally require object association. The external evaluation above specifies the transformation and reference used for each method.

\paragraph{Interpretation of the expanded method set.}
Tripo/Hunyuan provide the two-domain dense correspondence comparison. ShapeR, Amodal3R, SceneGen, and EFM3D span multiview object reconstruction, amodal completion, scene generation, and egocentric reconstruction. Their measured surface and screening results use the references and placements specified above. The output galleries show their native geometry separately from calibrated quality comparisons.

\section{Reference validity, categories, and reproducibility}
\paragraph{Independent-reference status.}
The dense support and scan-reference rendering share geometry. Dense results therefore measure scan-conditioned visible agreement. The external point references are capture-derived or associated with catalog geometry. Independent manual-mask agreement and inter-annotator reliability are not evaluated.

\paragraph{Category coverage.}
The visual inspection includes named instances such as chairs, cabinets, benches, kettles, guitars, baskets, candles, and airplanes across the source cohorts. These names support object-level exemplars, while category rates require a verified source taxonomy joined to the frozen IDs. The current evaluation therefore reports domain- and route-level results, with category-level accuracy reserved for a taxonomy-backed analysis.

\paragraph{Reproducibility.}
The evaluation preserves observation splits, candidate transforms, per-view measurements, and fixed decisions. Hosted-generation comparisons concern the evaluated outputs; variability across repeated provider calls is outside this study.

\paragraph{Data access and redistribution.}
Source datasets, pretrained models, and hosted assets retain their access and redistribution terms. Evaluation code specifies measurement, registration, splitting, and aggregation; obtaining source data and weights follows the respective providers' terms.

\clearpage
\onecolumn
\section{Registration and asset-choice evidence}
\label{sec:svisual-selection}
\subsection{Scene-level variation}
\begin{figure}[!ht]
\centering
\includegraphics[width=\textwidth]{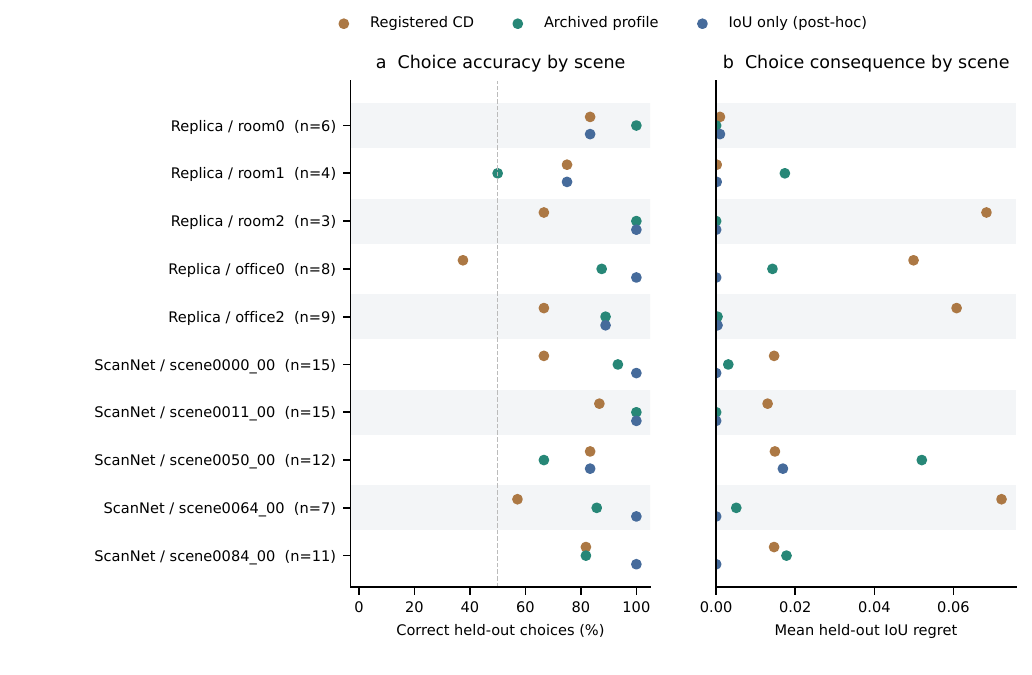}
\caption{\textbf{Asset-choice gains vary across scenes.} Every row contains the same paired instances for registered CD, the archived observation profile, and the post-hoc IoU-only rule. Panel (a) reports correct choices; panel (b) reports mean held-out IoU regret with zero assigned to failed views before aggregation. The dashed line is the two-candidate uniform-random accuracy. Counts are paired instances, and points are descriptive estimates without confidence intervals. The ten scene groups partition the 90-instance main comparison.}
\label{fig:ssceneselection}
\end{figure}

The paired cohort contains examples from all ten source scenes. The archived profile improves choice accuracy over registered CD in seven scenes, ties in one, and loses in two: Replica room1 and ScanNet scene0050\_00. Scene0084\_00 has equal accuracy but higher profile regret. \figref{fig:ssceneselection} shows both outcome measures on the same candidates, including these reversals. Scene sizes range from three to fifteen paired instances; the plotted values describe these groups and do not constitute a leave-one-scene-out predictor evaluation.

\clearpage
\subsection{Registration transfer and aggregate choice outcomes}
\begin{figure}[!ht]
\centering
\includegraphics[width=\textwidth]{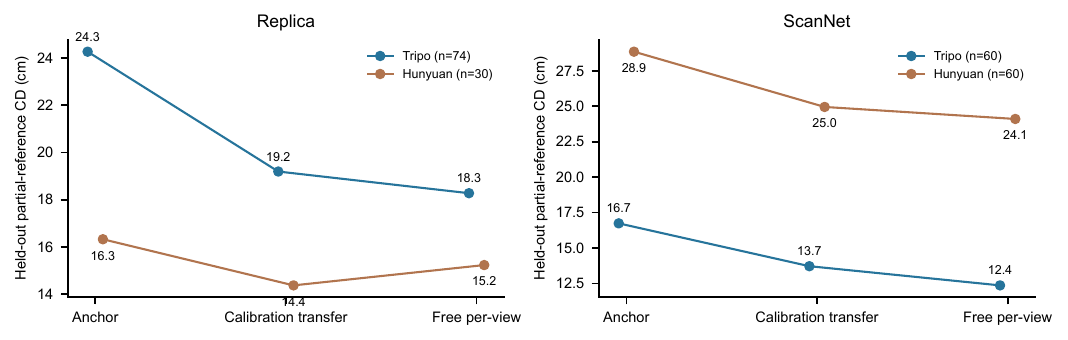}
\caption{\textbf{Registration affects held-out geometry as well as calibration geometry.} The candidate and world-metre reference conventions match the main table, but only held-out views enter this plot. Lines connect descriptive instance-macro means, not independent repeated trials; no confidence interval is implied. Free per-view fitting reads the target geometry and remains a sensitivity endpoint, not a selection feature.}
\end{figure}

\begin{figure}[!ht]
\centering
\includegraphics[width=\textwidth]{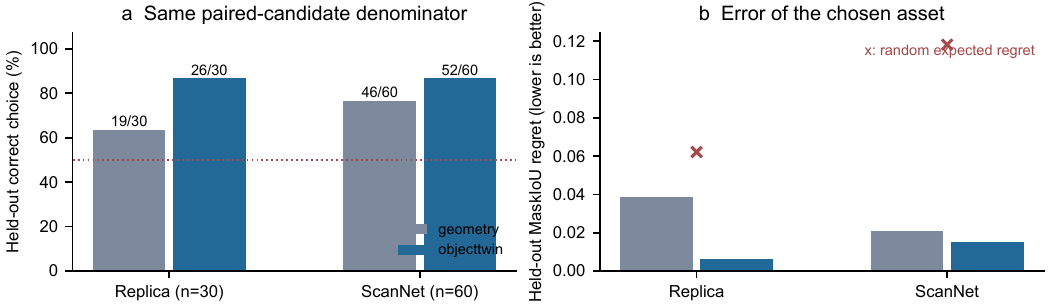}
\caption{\textbf{Magnitude of the paired-choice effect.} Correct choices and held-out regret use the same 30 Replica and 60 ScanNet instances as the main discordance figure. Missing held-out views are rescored as zero before averaging, while calibration choices remain unchanged. The dashed 50\% line and crosses are the uniform-random expectation in this two-candidate, non-tied cohort. Bars are point estimates without borrowed uncertainty intervals.}
\end{figure}

\begin{figure}[!ht]
\centering
\includegraphics[width=\textwidth,height=0.84\textheight,keepaspectratio]{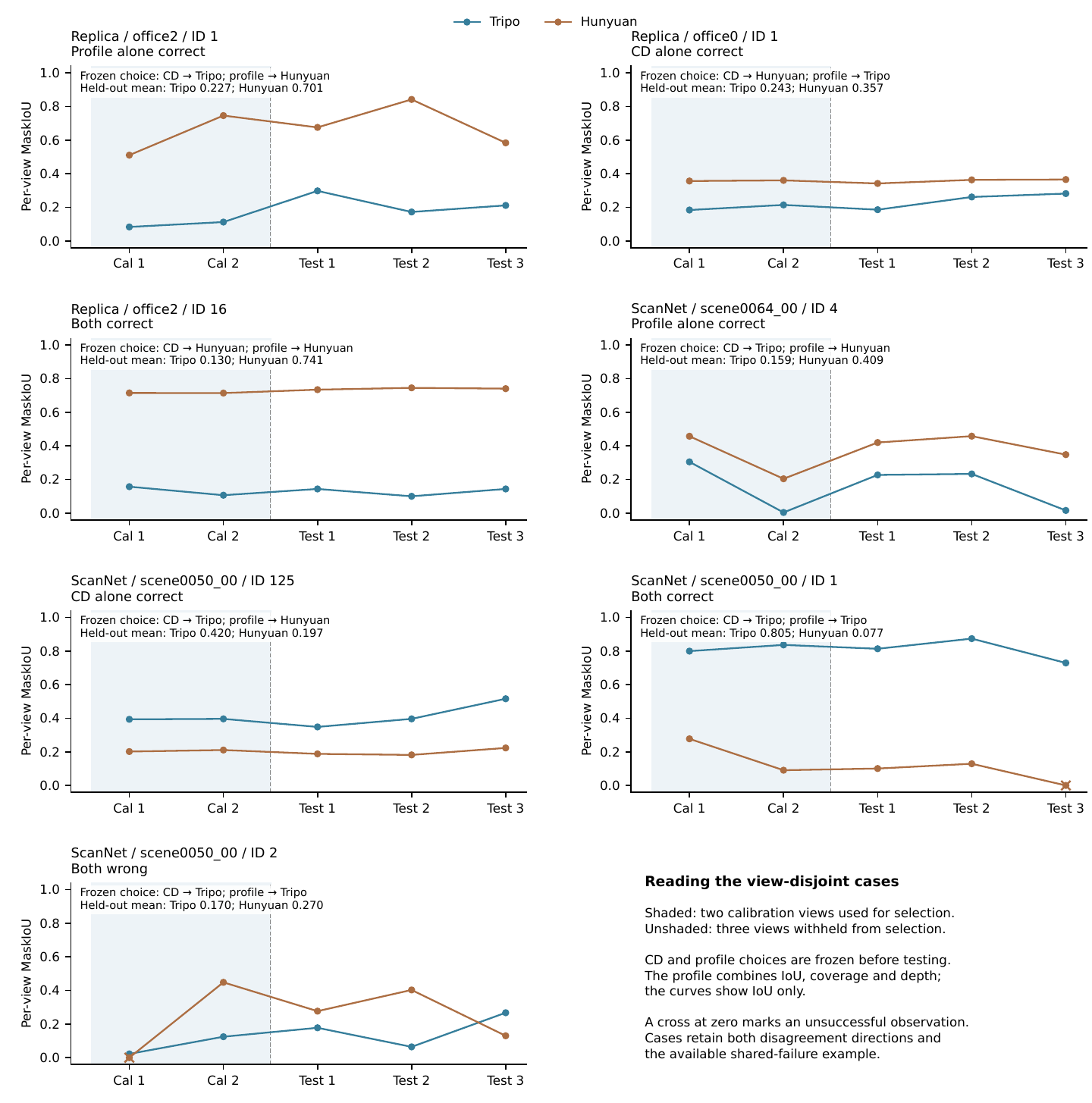}
\caption{\textbf{View-level evidence for all seven outcome-stratified cases.} Each panel retains two calibration and three held-out views for both frozen candidates. The shaded region identifies selection evidence; the dashed boundary separates views withheld from the choice. Lines connect ordered views, not temporal trials or repeated generations. Failed observations contribute zero and retain cross markers. The curves display IoU; the archived profile choice combines IoU, coverage, and depth. Cases include both directions of selector disagreement, common successes, and the common failure, under the same display rule as the preceding table. Exact view IDs and statuses accompany the source data.}
\label{fig:smultiviewdecisions}
\end{figure}

The fixed-state image comparisons show missing support and projected-shape errors with per-view scores. The decision-case table instead averages held-out views; \figref{fig:smultiviewdecisions} shows all calibration and test-view scores for those examples.

\clearpage
\subsection{Calibration and held-out views of the decision cases}

\clearpage
\section{Candidate provenance and fixed-state visual evidence}
\subsection{Historical production pipeline}
The following pipeline explains how the fixed Tripo/Hunyuan candidates were produced. The comparison evaluates those existing assets using the same captured object identities and calibrated cameras. The visual examples retain the original observations and candidate renders.
\begin{figure}[!ht]
\centering
\includegraphics[width=\textwidth]{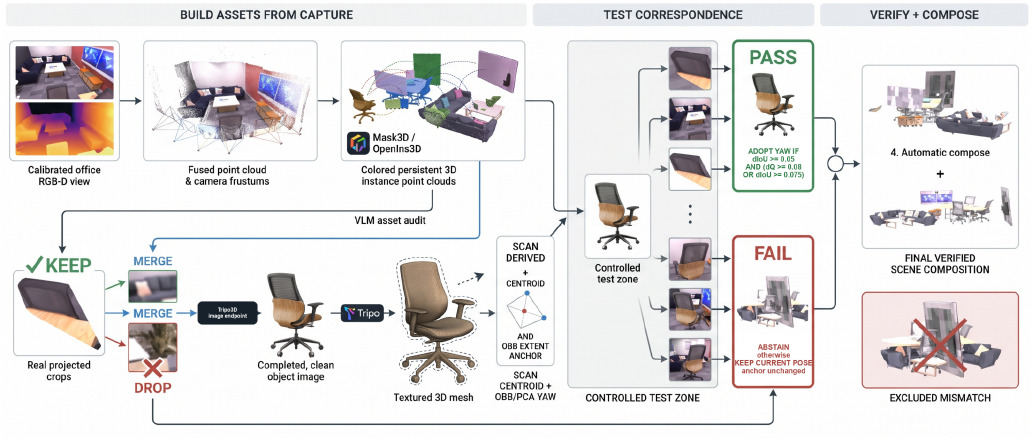}
\caption{\textbf{Complete historical asset-production architecture, preserved from the original study.} Capture fusion, persistent instances, crop filtering/completion, mesh generation, scan anchoring, pose-update testing, and scene composition retain their original relationships. The PASS/FAIL gates and thresholds concern the historical pose-update policy, not the new two-candidate selection rule or a validated admission threshold. This schematic explains candidate provenance; it is not a cross-method accuracy comparison.}
\label{fig:shistorical}
\end{figure}
\begin{figure}[!ht]
\centering
\includegraphics[width=\textwidth]{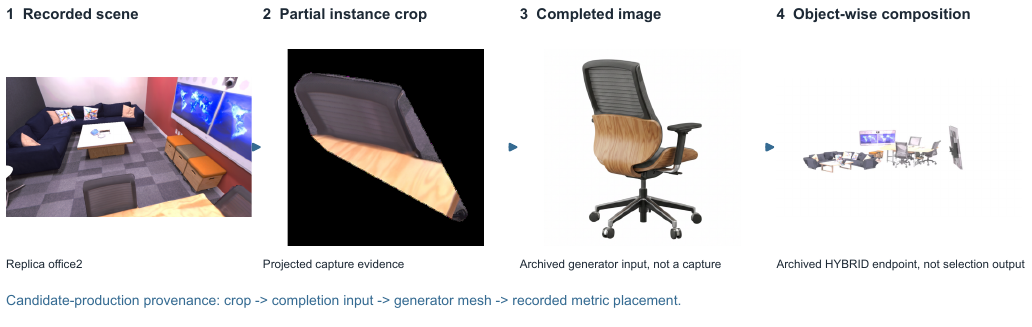}
\caption{\textbf{Recorded candidate-production artifacts.} The historical object-wise pipeline produces generator-ready inputs and anchored assets. The completed image is explicitly a generator input, not an observed image or a rendered mesh. The composition is an archived HYBRID endpoint, not an output from the new asset-selection experiment. This chain explains the Tripo/Hunyuan candidate provenance; it is not the proposed measurement interface.}
\label{fig:sproduction}
\end{figure}
\clearpage
\subsection{Complete archived fixed-state atlas}
Four six-case plates preserve all 24 archived examples with both scene context and object detail. Each case has a full-camera row followed by the same rectangular crop in six columns: RGB, scan-derived reference, Tripo, Hunyuan, and both error maps. The crop encloses the reference and both candidate supports, retaining all unsupported projection; its yellow boundary appears on the context RGB. The displayed scores belong to the shown view. No score is recomputed from a crop, and crop coordinates and source checksums accompany every case.

\begin{figure}[!ht]
\centering
\includegraphics[width=\textwidth,height=0.74\textheight,keepaspectratio]{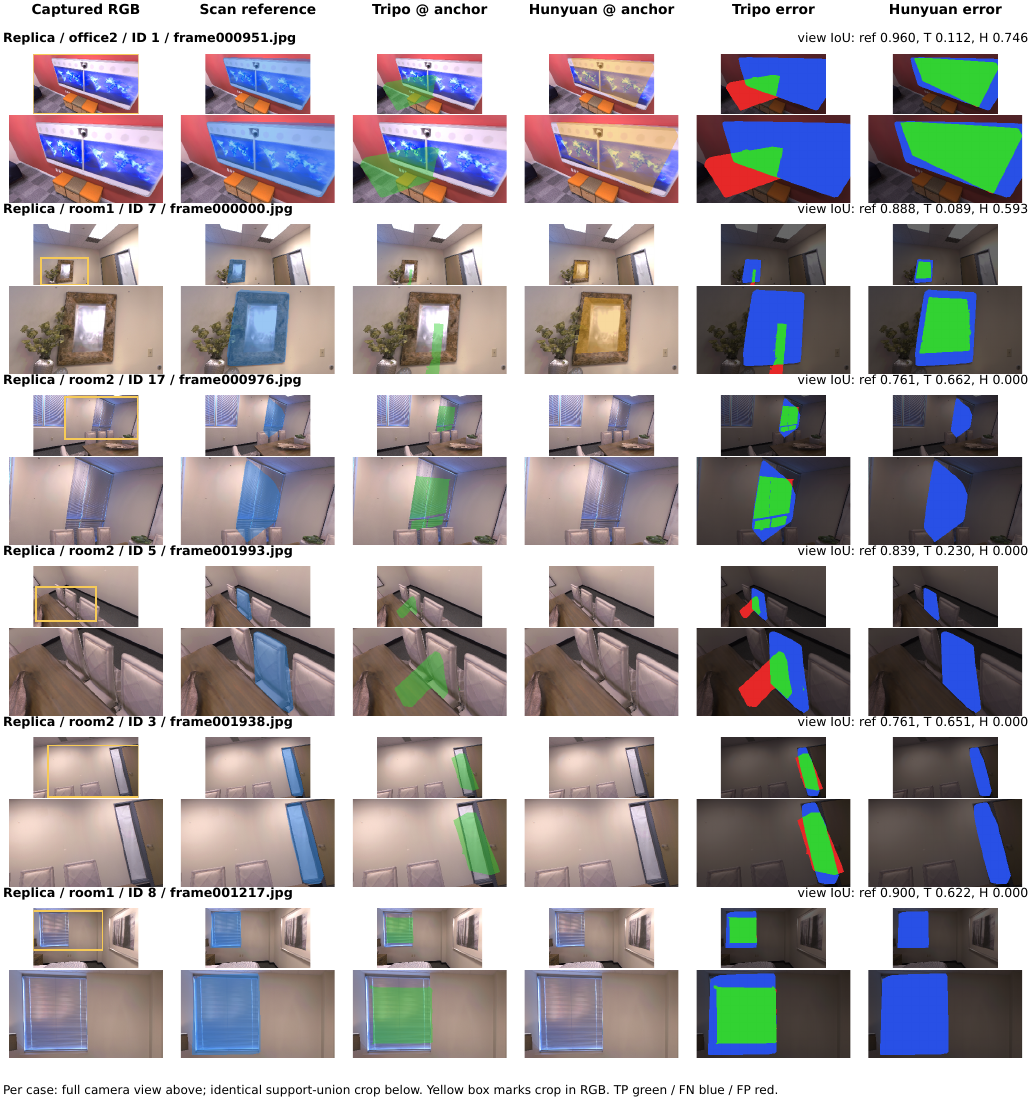}
\caption{\textbf{Replica atlas, cases 1--6 of 12.} Full-camera views and matched detail rows expose support errors at two scales. Six evidence columns preserve the reference, both frozen candidates, and their TP/FN/FP maps. Instance/view identities and per-view scores identify each case. All columns retain the archived camera and placement.}
\end{figure}

\clearpage
\begin{figure}[!ht]
\centering
\includegraphics[width=\textwidth,height=0.87\textheight,keepaspectratio]{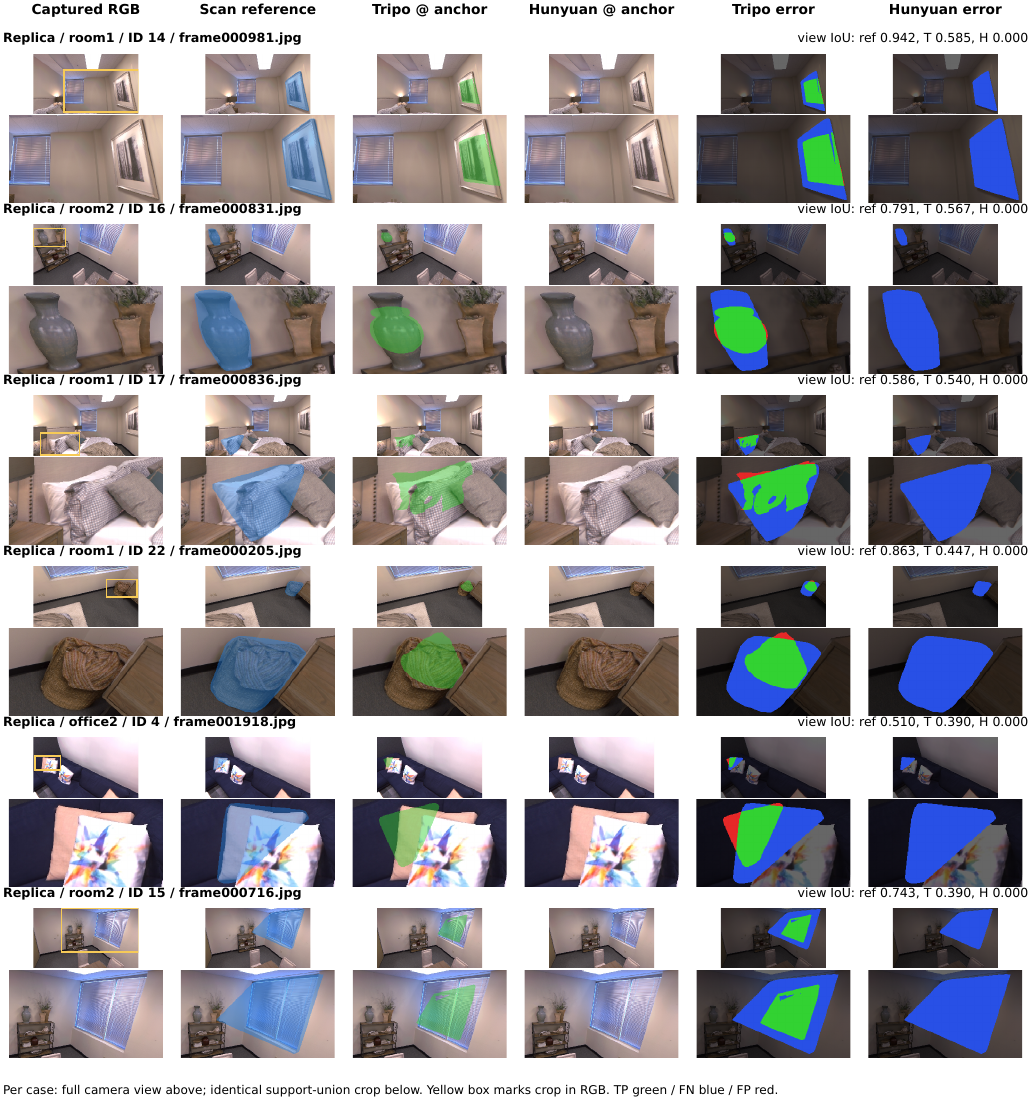}
\caption{\textbf{Replica atlas, cases 7--12 of 12.} Paired context/detail rows reveal small-object and boundary errors while retaining the original view. Identical support-union crops apply to all six columns of each case. Together the two Replica plates show every archived example from three scenes; the numerical evaluation covers five Replica scenes.}
\end{figure}

\clearpage
\begin{figure}[!ht]
\centering
\includegraphics[width=\textwidth,height=0.87\textheight,keepaspectratio]{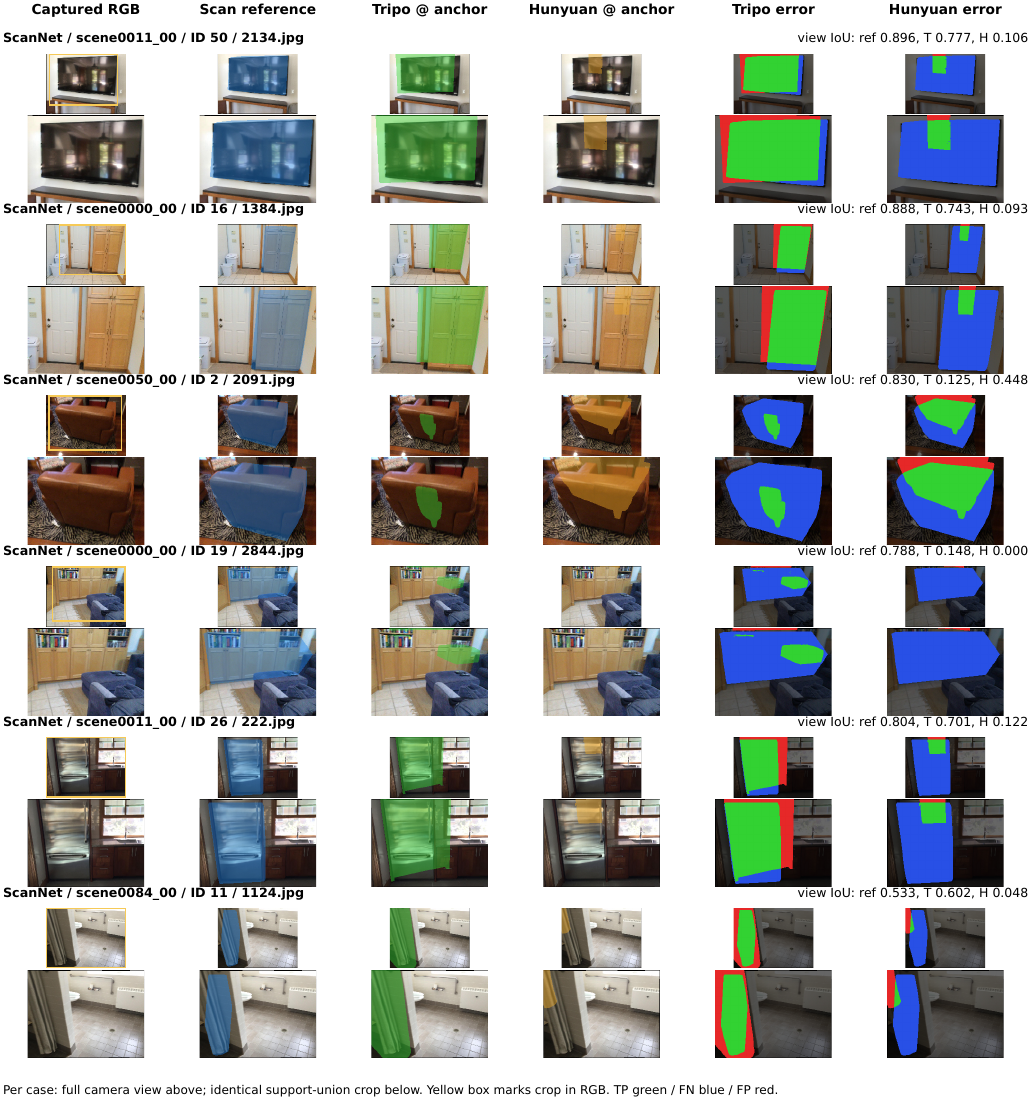}
\caption{\textbf{ScanNet atlas, cases 1--6 of 12.} Context/detail pairs retain capture noise and occlusion under the same six-column comparison. Each case uses one archived camera and its exact view scores. These are original fixed-state observations; the decision study's view-disjoint cases appear separately in \figref{fig:smultiviewdecisions}.}
\end{figure}

\clearpage
\begin{figure}[!ht]
\centering
\includegraphics[width=\textwidth,height=0.87\textheight,keepaspectratio]{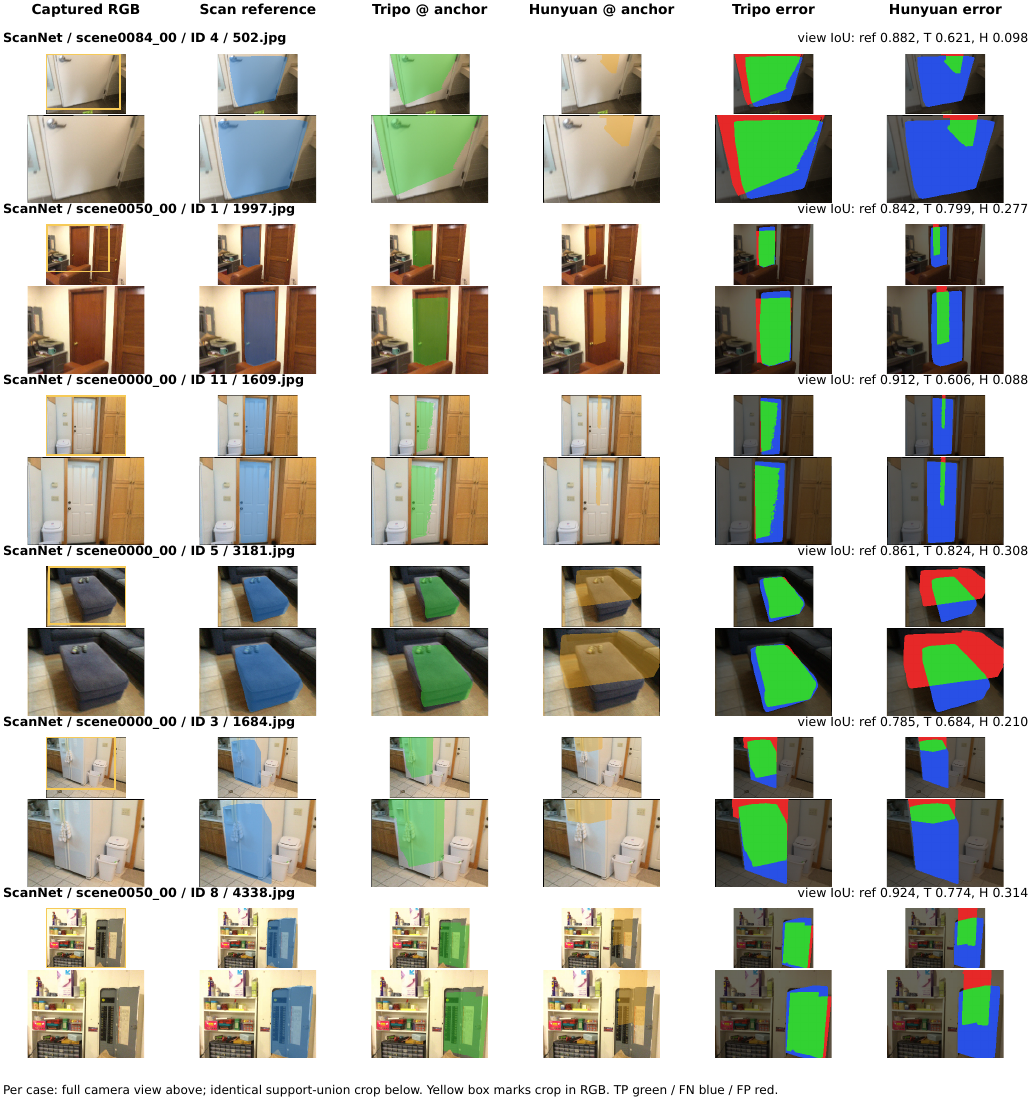}
\caption{\textbf{ScanNet atlas, cases 7--12 of 12.} These context/detail pairs complete the original ScanNet qualitative inventory. The 12 illustrated instances span four scenes, while quantitative results cover five. Support-union crops retain both missing reference support and unsupported candidate projection.}
\end{figure}

\clearpage
\section{Reconstruction examples in native coordinates}
Seven object examples cover all seven source scene groups in the ShapeR cohort. The DTC examples compare active and passive captures for three catalog objects, including an unavailable EFM3D output. Two orthographic views show each output with uniform shading and independent centering and scale. These displays illustrate the predicted geometry; calibrated correspondence is evaluated by the surface and screening comparisons above.

\begin{figure}[!ht]
\centering
\includegraphics[width=\textwidth,height=0.78\textheight,keepaspectratio]{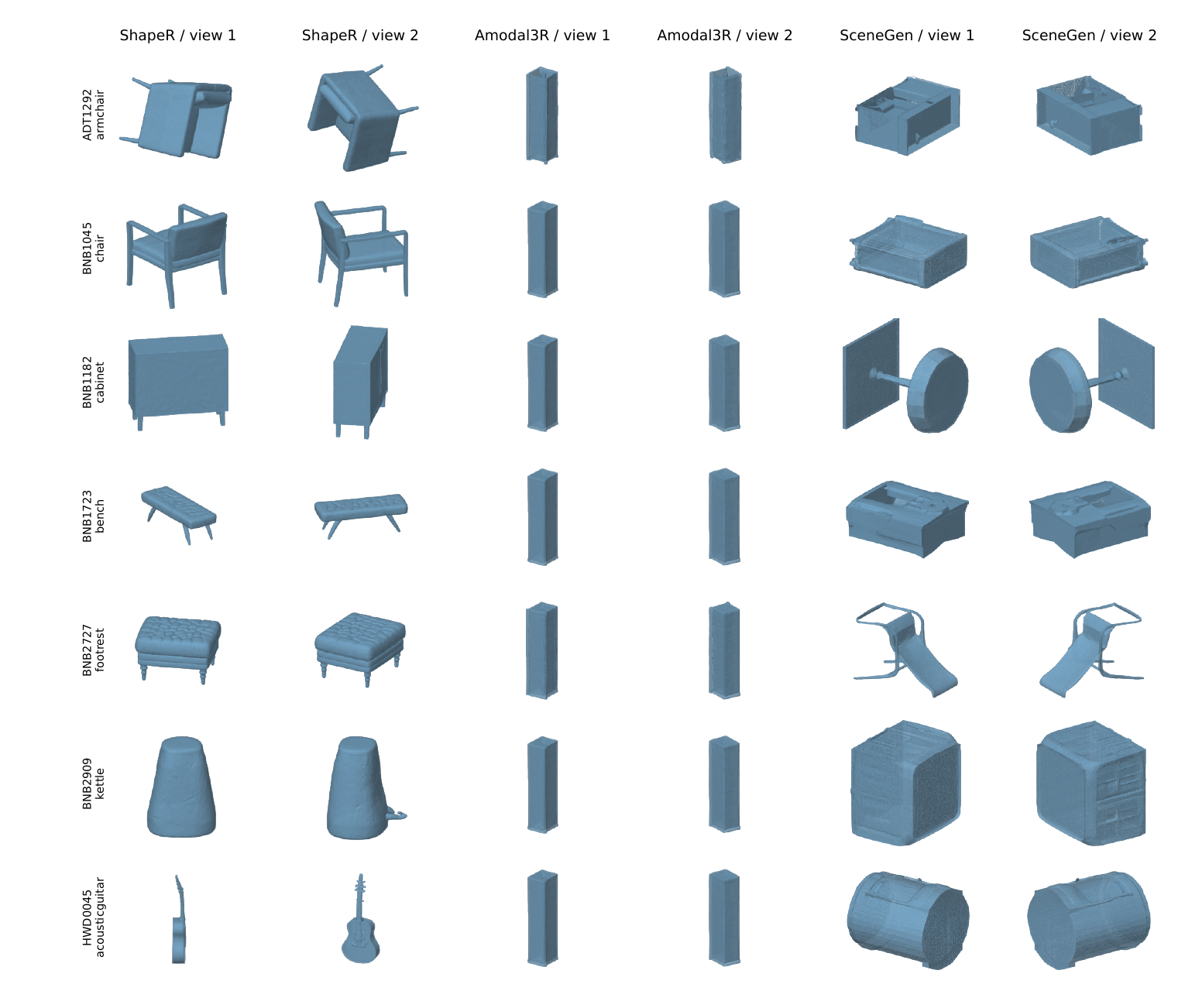}
\caption{\textbf{Reconstruction examples from seven source scenes.} Each row shows one object reconstructed by three pretrained methods, with two views per output. Independent display normalization preserves shape detail while removing metric size and placement from this qualitative comparison. Numerical surface quality uses the calibrated observations described above.}
\label{fig:snativegallery}
\end{figure}

\clearpage

\end{document}